%% file: main.tex
\documentclass[10pt,twocolumn,letterpaper]{article}

 \usepackage{cvpr}              
\usepackage{adjustbox}
\usepackage{booktabs}

\usepackage{graphicx}
\usepackage{amsmath}
\usepackage{amssymb}
\usepackage{booktabs}

\usepackage{amsthm, color, listings, comment}

\usepackage{bbm}
\usepackage{dsfont}

\usepackage[utf8]{inputenc} 
\usepackage[T1]{fontenc}    
\usepackage{url}            
\usepackage{booktabs}       
\usepackage{amsfonts}       
\usepackage{nicefrac}       
\usepackage{microtype}      
\newtheorem{definition}{Definition}
\usepackage{graphicx}
\usepackage{amsmath}
\usepackage{amssymb}
\usepackage{microtype}
\usepackage{graphicx}
\usepackage{booktabs} 
\usepackage{lipsum}
\usepackage{amsmath}
\usepackage{adjustbox}
\usepackage{algorithm}
\usepackage{algorithmic}
\usepackage{graphicx}
\usepackage{enumitem}
\usepackage{pifont}
\usepackage{wrapfig}
\usepackage{bm}
\usepackage{float}
\usepackage[accsupp]{axessibility}

\newtheorem{assumption}{Assumption}
\newtheorem{theorem}{Theorem}

\newtheorem{proposition}{Proposition}
\input{preamble}

\definecolor{cvprblue}{rgb}{0.21,0.49,0.74}

\def\paperID{24} 
\def\confName{CVPR}
\def\confYear{2024}

\title{Data-Efficient and Robust Task Selection for Meta-Learning}

\author{Donglin Zhan\\
Columbia University\\
New York, NY\\
{\tt\small donglin.zhan@columbia.edu}
\and
James Anderson\\
Columbia University\\
New York, NY\\
{\tt\small james.anderson@columbia.edu}
}

\begin{document}
\maketitle

\begin{abstract}
Meta-learning methods typically learn tasks under the assumption that all tasks are equally important. However, this assumption is often not valid. In real-world applications, tasks can vary both in their importance during different training stages and in whether they contain noisy labeled data or not, making a uniform approach suboptimal. To address these issues, we propose the Data-Efficient and Robust Task Selection (DERTS) algorithm, which can be incorporated into both gradient and metric-based meta-learning algorithms. DERTS selects weighted subsets of tasks from task pools by minimizing the approximation error of the full gradient of task pools in the meta-training stage. The selected tasks are efficient for rapid training and robust towards noisy label scenarios. Unlike existing algorithms, DERTS does not require any architecture modification for training and can handle noisy label data in both the support and query sets. Analysis of DERTS shows that the algorithm follows similar training dynamics as learning on the full task pools. Experiments show that DERTS outperforms existing sampling strategies for meta-learning on both gradient-based and metric-based meta-learning algorithms in limited data budget and noisy task settings.

\end{abstract}

\section{Introduction}

\vspace{-0.2cm}

Meta-learning methods have been extensively studied and applied in computer vision, natural language processing, and robotics \citep{hospedales2021meta}. The key idea of meta-learning is to mimic the few-shot situations faced at test time by randomly sampling classes in meta-training data to construct tasks for episodic training. Most existing meta-learning methods randomly sample meta-training tasks with a uniform probability  \citep{finn2017model,snell2017prototypical,raghu2020rapid}. The assumption behind such uniform sampling is that all tasks are contribute equally. In real-world scenarios, things often differ. Considering that the importance of tasks can change during the training process, the diversity of tasks, and the probability of mislabeled data within tasks, some tasks might be more informative for the meta-training process than others. In this paper, we specifically focus on two scenarios: a limited data budget and a noisy task setting.

First, making the meta-training process more efficient is an essential issue, particularly in scenarios with a limited budget of data (e.g., the number of tasks and data learned in the meta-training process). Current meta-learning algorithms require many tasks for meta-training episodes, which may contain redundant information.

In this work,  we raise the natural concern that not all the tasks are created equal. A coarse-grained task that contains the classification of ``Dog" and ``Laptop" is much easier to learn for the meta-model than the fine-grained task that includes a more difficult classification (e.g., ``Dog" or ``Cat") \citep{liu2020adaptive}. In meta-training episodes, selecting the most benignly informative subset of tasks could benefit this process by decreasing the computational load.

Second, most meta-learning methods presuppose that the few support set training samples were chosen correctly to represent their class \citep{finn2017model,snell2017prototypical,raghu2020rapid}. Unfortunately, such assurances are not usually provided in real-world scenarios. In reality, mislabeled samples are frequently present in even highly annotated and maintained datasets \citep{natarajan2013learning,li2017learning,han2019deep,song2022learning} as a result of automated weakly supervised annotation, ambiguity, or even human error. Suppose some tasks with noisy labels are fed to the meta-training process. In that case, the noisy-labeled data in the support set will hinder the adaptation step of the meta-training process, which can result in invalid adaptation and further make the base model incorrect on query evaluation. Furthermore, if noisy data is also in the query set, the misleading gradient would be updated for the meta-model, which hinders the generalization capacity of the meta-model during the meta-testing stage. Therefore, developing methods to avoid training meta-models on heavily label-corrupted tasks is essential for the setting with noisy data as well.

The two aspects mentioned above still need to be thoroughly studied: (i) For limited computational and data budgets, existing approaches construct information-theoretic criteria for evaluating and selecting task \citep{kaddour2020probabilistic,li2020difficulty,luna2020information} or building extra module for learning the sampling probability \citep{yao2021meta} with high computation cost. (ii) Previous work that considers meta-learning with noisy data \citep{mazumder2021rnnp,yao2021meta,liang2022few} make the assumption that noisy data only exist in the support set of tasks. This assumption may be unrealistic for applications under the ``uncertain noise'' setting (e.g., tasks with different noise ratios, noisy data could be in both support sets and query sets, and identities are not exposed to models).

We propose the \textit{Data-Efficient and Robust Task Selection (DERTS)} algorithm for meta-learning that can select appropriate tasks in the meta-training stage with efficiency for rapid training \emph{and} robustness towards noisy data scenarios inspired by \citep{mirzasoleiman2020coresets2,balakrishnan2022diverse}. Unlike existing works for task sampling, DERTS does not require any architecture modification; it only requires an iterative task pool to store tasks for the model to select. 

DERTS selects weighted subsets of tasks from task pools by minimizing the approximation error of the full gradient of task pools in the meta-training stage.  Furthermore, by dropping tasks in the subset with potentially high estimated gradient norms, we find the proposed algorithm is robust toward the noisy task scenario.

\textbf{Contributions:} We propose a data-efficient and robust task selection algorithm for meta-learning in limited data budget and noisy label settings.

\begin{itemize}
[leftmargin = *]
    \item We formulate a weighted subset selection objective that minimizes the approximation error of the full gradient on episodic task pools. Due to the submodularity of the approximation error, we apply a stochastic greedy approximation to the solution of the derived optimization objective. This method can be easily incorporated into both gradient-based and metric-based meta-learning schemes.

    \item  We extend the selection algorithm to a challenging noisy task setting with mislabeled data in both the support and query sets by dropping tasks with a large estimated gradient norm.

    \item We provide a theoretical analysis that proves that learning on a subset of tasks produces similar training dynamics than if trained on the full task pool. We provide an upper-bound the difference between the model trained with our task selection algorithm and the model trained with all the meta-training tasks. Our result highlights a fundamental bias when applying \emph{any} selection method.

    \item Extensive experiments show that DERTS outperforms the state-of-the-art sampling strategies for meta-learning with both gradient-based and metric-based meta-learning algorithms on a limited budget and noisy task settings. The performance improvement averages between 3\% and 5\%, while also achieving a speedup of more than three times.

\end{itemize}

\vspace{-0.2cm}

\section{Related Work}

\vspace{-0.1cm}

Meta-Learning focuses on rapidly adapting to new unseen tasks by learning prior knowledge through training on many similar tasks. Metric-based methods classify query examples based on their similarity to each class's support data, learning a transferable embedding space for evaluation and prediction \citep{snell2017prototypical,garcia2017few,sung2018learning,oreshkin2018tadam,ye2020few,yang2020context,liang2022few,yang2022efficient}. Other than metric-based approaches, optimization-based methods fine-tune model parameters on a few support examples \citep{finn2017model,antoniou2018train,nichol2018first,raghu2020rapid,sun2021towards,wang2022learning,chi2022metafscil,liu2022few,wang2022meta,yu2023enhancing}. However, most previous work assumes tasks contribute equally to the meta-training stage. Recently, some work has focused on optimizing the sampling distribution strategy for meta-learning \citep{liu2020adaptive,kaddour2020probabilistic,luna2020information}. 

Arnold et al. \citep{arnold2021uniform} apply a uniform episodic sampler (US) to reweight tasks based on the observation that the task learning difficulty follows a normal distribution for arbitrary meta-datasets and model architectures. Although promising, the empirical assertion that US is based on may face challenges when encountering tasks with disturbances and uncertainties. Yao et al. \citep{yao2021meta} propose an adaptive task scheduler (ATS) to jointly learn the sampling probability for tasks in the candidate pool to address the noisy label issue. Specifically, ATS takes the loss value and inner product of the gradient on the support set and query set and outputs the corresponding sampling probability. The robustness of ATS is based on the assumption that the noise only exists in the support set. When breaking this condition, ATS is likely to be less robust in the noisy task setting. Unlike US and ATS, our algorithm, DERTS, focuses on both data-efficiency \emph{and} robustness. For the data-efficient perspective, DERTS aims to approximate the performance of full training episodes via only training on a subset of tasks. As for the robustness issue, we allow for the fact that the noisy data can be present in  the support set \emph{and} query set without any side information; this poses significant challenges in the evaluation of robustness.

With respect to data selection and sampling aspects of the problem, there have been efforts to take advantage of the difference in importance among various samples to reduce the variance and improve the convergence rate of stochastic optimization methods. Those that apply to overparameterized models employ either the gradient norm \citep{katharopoulos2018not} or the loss function \citep{loshchilov2015online,schaul2015prioritized} to compute each sample's importance. Recently, a series of data selection strategies have discussed selecting subsets of the data for efficient training. CRAIG \citep{mirzasoleiman2020coresets1} finds the subsets by greedily maximizing a submodular function and provides convergence guarantees to a neighborhood of the optimal solution for both convex and non-convex models. GRADMATCH \citep{killamsetty2021retrieve} proposes a variation to address the same objective using orthogonal matching pursuit. CREST \citep{yang2023towards} models the non-convex loss as a series of quadratic functions for extracting subset. However, the work mentioned above focuses on the standard supervised learning setting and it is not clear how such approaches can be ported over to task selection for episodic training in meta-learning due to their formulation as a bilevel-optimization problem.

\begin{figure*}
     \centering
     \vspace{-0.3cm}
    \includegraphics[scale=0.52]{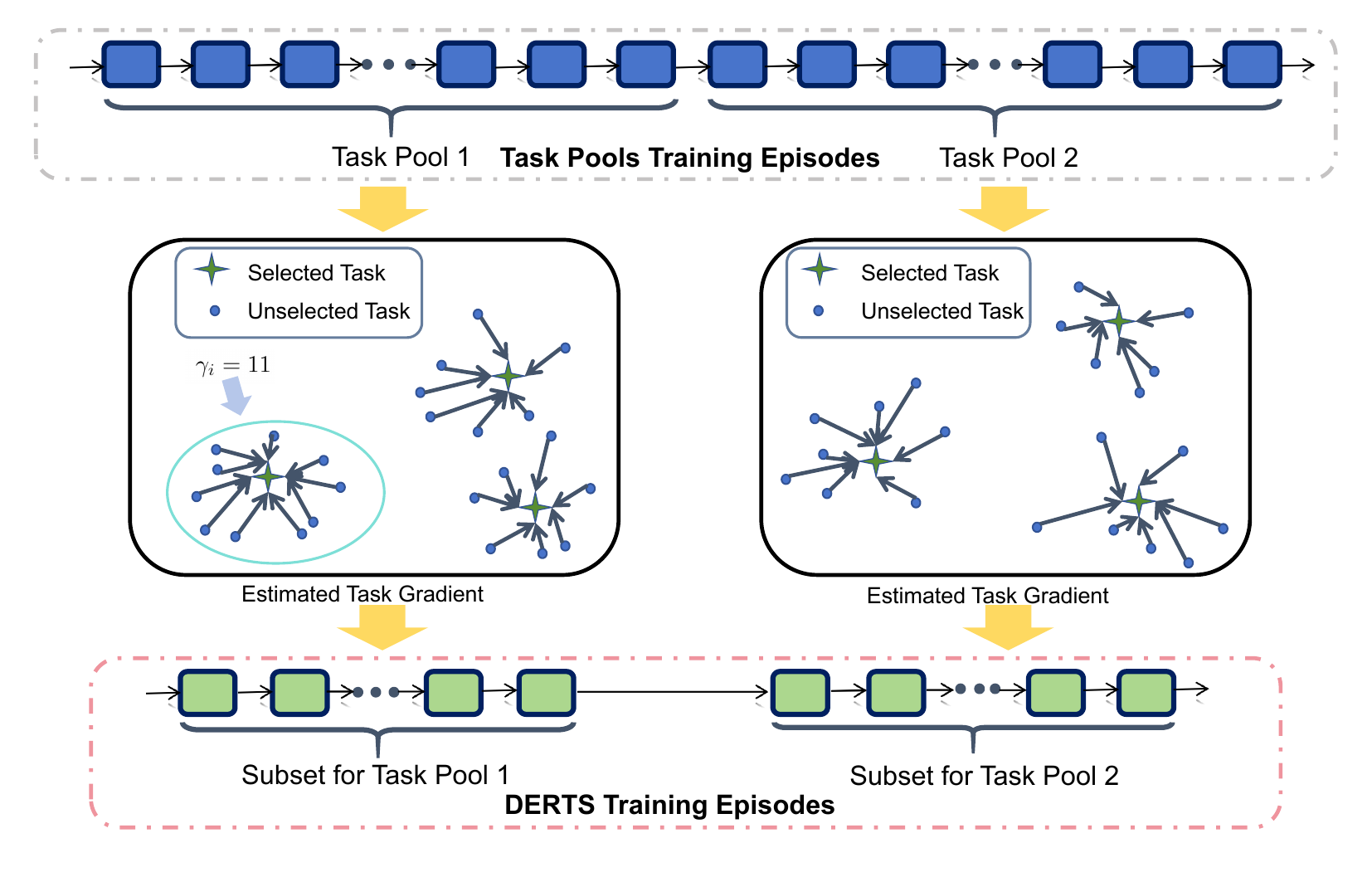}
     \vspace{-0.2cm}
     \caption{ DERTS requires task pools to store episodic tasks sampled from task distributions. With the efficient gradient estimation in sec.\ref{sub2}, the gradients of all the tasks stored in the task pool are computed. According to the approximation formulated in sec. \ref{sub1} and optimization objective in sec. \ref{sub2}, a subset of tasks with corresponding weights is constructed to approximate the task pool gradient. The meta-model then conducts a training process on the subsets instead of task pools.}
    \label{DERTS}
    \vspace{-0.5cm}
\end{figure*}


\vspace{-0.2cm}

\section{Background} 
\vspace{-0.2cm}
We consider the standard meta-learning setting, where given a set of training tasks $\mathcal{T}_1, \ldots, \mathcal{T}_n$ sampled from task distribution $p(\mathcal{T})$, we would like to learn a good parameter initialization $\theta^{*}$ for a predictive model $f_\theta$ such that it can be quickly adapted to new tasks given only a limited amount of data (i.e., few-shot regime). Each task $\mathcal{T}_i$ has a support set of labeled data $\mathcal{D}_i^s=\left\{\mathbf{X}_i^s, \mathbf{Y}_i^s\right\}=\left\{\left(\mathbf{x}_{i, j}^s, \mathbf{y}_{i, j}^s\right)\right\}_{j=1}^{N^s}$ and a query set, $\mathcal{D}_i^q=\left\{\mathbf{X}_i^q, \mathbf{Y}_i^q\right\}=\left\{\left(\mathbf{x}_{i, j}^q, \mathbf{y}_{i, j}^q\right)\right\}_{j=1}^{N^q}$ of labeled data, where $N_s$ and $N_q$ refer to the size of support and query sets respectively. Given the predictive model  $f_{\theta}$, meta-learning algorithms first train the base model on meta-training tasks. Then, during the meta-testing stage, the well-trained base model is applied to the new task $\mathcal{T}_t$ by taking a few adaptation steps on its support set $\mathcal{D}_t^s$. Finally, the performance is evaluated on the query set $\mathcal{D}_t^q$. We provide a brief introduction to gradient-based and metric-based algorithms in \textbf{Appendix A}.

\vspace{-0.3cm}

\section{Efficient and Robust Task Selection}

\vspace{-0.1cm}

The conceptual idea behind DERTS is to (i) select subsets of tasks from the overall task pools, (ii) assign a weight to each task in the subset that captures their relative importance, and (iii) use the tasks in the subset for meta-training with corresponding weights. We first formulate a full gradient approximation for the task pools in sec.~\ref{sub1}. Secondly, in sec.~\ref{sub2}, we establish an optimization objective for task selection and provide the solution based on submodular maximization. In addition, we provide a modified DERTS to address the scenario with noisy label tasks in sec.~\ref{sub4}. We also provide a theoretical analysis of DERTS in sec.~\ref{sub5}. \textbf{Figure~\ref{DERTS}} demonstrates the workflow of DERTS. We also note that DERTS can be easily incorporated into widely-used gradient-based and metric-based meta-learning schemes.

\subsection{Full Gradient Approximation for Episodic Task Pools} \label{sub1}

\vspace{-0.1cm}

We start by selecting  a  sample  of candidate tasks drawn from the task distribution $p(\mathcal{T})$ in advance and store them in a task pool $\mathcal{M}$. Suppose for a  task pool $\mathcal{M}:= \left\{\mathcal{T}_{j} | j = 1,2,\ldots, N\right\}$, we want to select a subset of tasks $\mathcal{S}: = \left\{\mathcal{T}_{i} ~|~ i = \alpha_1,\alpha_2,\ldots,\alpha_k\right\}$, where $\alpha_i \in [N]$ and $k < N$, with corresponding weights $\left\{\gamma_{i} ~|~ i = 1,2,\ldots,k\right\}$ such that the gradient for training on $\mathcal S$ approximates the gradient on $\mathcal{M}$. We now describe our approach for determining the weights:   Let $\Gamma: \mathcal{M} \rightarrow \mathcal{S}$ be a mapping from the task pool $\mathcal{M}$ to the subset $\mathcal{S}$ that maps a task $\mathcal{T}_{j}$ from $\mathcal{M}$ to a task $\mathcal{T}_{i}$ in $\mathcal{S}$. For simplicity, we denote $\Gamma(\mathcal{T}_{j}) = \mathcal{T}_{i}$ as $\Gamma(j) =i$ and $\mathcal{T}_j \in \mathcal{M}$ as $j \in \mathcal{M}$. Let $\mathcal{S}_\mathcal{M}^C$ denote the complement of $\mathcal{S}$ in $\mathcal{M}$.  Similar to \cite{balakrishnan2022diverse}, we define the weight $\gamma_i$ of the selected task $\mathcal{T}_{i} \in \mathcal{S}$ as
\vspace{-0.1cm}
\begin{equation}
\label{est.1}
\gamma_i := \sum_{j \in \mathcal{M}}\mathbbm{1}_{\{\Gamma(j)=i\}}  = \sum_{j \in \mathcal{S}} \mathbbm{1}_{\{ \Gamma(j) = i\} } + \sum_{j \in \mathcal{S}^C_\mathcal{M}}\mathbbm{1}_{\{\Gamma(j)=i\}},
\end{equation}
where $1_\mathcal C$ is the indicator function for the set $\mathcal C$.

The first term on the RHS of eq.~\eqref{est.1} is equal to 1 since $\Gamma$ is identity in $\mathcal{S}$. For the second term, by taking the summation over $e \in \mathcal{S}^C_M$, we are calculating how many elements in $\mathcal{S}^C_M$ are mapped into task $\mathcal{T}_{i}$ in $\mathcal{S}$.

As we don't know $\mathcal S$ we cannot compute $\gamma_i$ directly from eq.~\eqref{est.1}. Instead we formulate an optimization problem for establishing how to select the set $\mathcal S$ from $\mathcal M$ in such a manner that when training is performed on the tasks in $\mathcal S$ the training dynamics (i.e., the gradients at each iteration)  approximate the gradients that would arise had we trained on the full task pool $\mathcal M$. As in the standard meta-training paradigm, we explore making the approximation of each task's gradient on the loss function of the task-adapted model $\phi_i$ (i.e., $\phi_i = \theta - \eta \nabla_{\theta}\mathcal{L}\left(f_{\theta} ; \mathcal{D}_{i}^s\right)$) on the query set $\mathcal{D}_{i}^q$ that is updated by the meta-model $\theta$. The gradient on task $\mathcal{T}_{i}$'s query is defined as
$
    \nabla_{\theta}\mathcal{L}\left(f_{\phi_{i}} ; \mathcal{D}_{i}^q\right) = \sum_{j=1}^{N^{q}}{\nabla_{\theta}\mathcal{L}\left(f_{{\phi}_{i}} ; \left(\mathbf{x}_{i, j}^q, \mathbf{y}_{i, j}^q\right) \right)}.
$
 Using the definition of the mapping function $\Gamma$ as defined above, DERTS uses the approximate gradient
 \vspace{-0.1cm}
\begin{equation*}
    \sum_{j \in \mathcal{M}}   \nabla_{\theta} \mathcal{L}\left(f_{\phi_{\Gamma(j)}} ; \mathcal{D}_{\Gamma(j)}^q\right)   
    = \sum_{j \in \mathcal{S}}\gamma_j \nabla_{\theta}\mathcal{L}\left(f_{\phi_{j}} ; \mathcal{D}_{j}^q\right),
\end{equation*}
which is taken over the set $\mathcal S$. In contrast, standard meta-learning algorithms use the full gradient on $\mathcal M$.

The gradient error  incurred by training on $\mathcal S$ instead of $\mathcal M$ is given by 
\vspace{-0.1cm}
\begin{equation}
\begin{split}
\label{approx}
\left\|\sum_{j \in \mathcal{M}} \nabla_{\theta}\mathcal{L}\left(f_{\phi_{j}} ; \mathcal{D}_{j}^q\right) - \sum_{j \in \mathcal{M}} \nabla_{\theta}\mathcal{L}\left(f_{\phi_{\Gamma(j)}} ; \mathcal{D}_{\Gamma(j)}^q\right)\right\| \\
= \left\| \sum_{j \in \mathcal{M}} \nabla_\theta \mathcal{L}\left(f_{\phi_{j}} ; \mathcal{D}_j^q\right)-\sum_{i \in \mathcal{S}} \gamma_{i} \nabla_{\theta}\mathcal{L}\left(f_{\phi_{i}} ; \mathcal{D}_{i}^q\right)\right\|.
\end{split}
\end{equation}
Ideally, we would like to make this error as small as possible. Unfortunately, we do not yet know $\Gamma$ and hence $S$, and so we cannot evaluate the error, and directly optimizing over $\mathcal S$ is NP-hard. However, we can upper bound the RHS of \eqref{approx} by a function that can be optimized over, specifically we have that

\vspace{-0.3cm}

\begin{equation}
\begin{split}
\left\| \sum_{j \in \mathcal{M}} \nabla_\theta \mathcal{L}\left(f_{\phi_{j}} ; \mathcal{D}_j^q\right)-\sum_{i \in \mathcal{S}} \gamma_{i} \nabla_{\theta}\mathcal{L}\left(f_{\phi_{i}} ; \mathcal{D}_{i}^q\right)\right\| \\
\leq 
\sum_{j \in \mathcal{M}} \min _{i \in \mathcal{S}}\left\| \nabla_{\theta}\mathcal{L}\left(f_{\phi_{j}} ; \mathcal{D}_{j}^q\right) -  \nabla_{\theta}\mathcal{L}\left(f_{\phi_{i}} ; \mathcal{D}_{i}^q\right)\right\|
\label{approx2}
\end{split}
\vspace{-0.3cm}
\end{equation}

which naturally leads to defining the approximation criteria as minimizing the RHS of~\eqref{approx2}. According to inequality~\eqref{approx2}, by assuming $\mathcal{S}$ is fixed, assigning the mapping $\Gamma$ to map the task in $\mathcal{M}$ to the closest element in $\mathcal{S}$ in the gradient space will minimize and upper bound on the gradient approximation error. Thus, $\gamma_{i}$ associated with mapping $\Gamma$ is computed as $$
    \gamma_i=\sum_{j \in \mathcal{M}} \mathbbm{1}_{\left\{j=\operatorname{argmin}_{\mathcal{T}_{i} \in \mathcal{S}}\left\|\nabla_\theta \mathcal{L}\left(f_\phi ; \mathcal{D}_i^q\right)-\nabla_\theta \mathcal{L}\left(f_\phi ; \mathcal{D}_j^q\right)\right\|\right\}} \text{.}$$

\vspace{-0.1cm}

\subsection{Extracting Subsets Efficiently} \label{sub2}
\vspace{-0.2cm}

Computing the explicit task gradient $\nabla_{\theta}\mathcal{L}\left(f_{\phi_{j}}; \mathcal{D}_{j}^q\right)$ that updates the meta-model is time-consuming and incurs a large computational cost. As shown in \cite{katharopoulos2018not}, the variation of the gradient norm is mainly captured by the gradient of the loss function with respect to the pre-activation outputs of the last layer. Suppose for a few-shot classification task, the above estimation only requires a forward computation on the last layer. E.g., for a softmax layer as the last, the gradients of the loss with respect to the input of the softmax layer for $\left(\mathbf{x}_{i, j}^q, \mathbf{y}_{i, j}^q\right)$ is $l_{i}$-$y_{i}$, where $l_{i}$ is the logits and $y_{i}$ is the encoded label. We extend the result of \cite{yang2023towards} to estimate the task-gradient $\nabla_{\theta}\mathcal{L}\left(f_{\phi_{i}}; \mathcal{D}_{i}^q\right)$ and denote it as $\Tilde{g}_i$. This approximation indicates the computation of gradient estimation $\Tilde{g}_{i}$ on task $\mathcal{T}_{i}$ is marginally more costly to compute than the value of the loss. We show the extended results and analysis in \textbf{Appendix B}.

Minimizing the RHS of~\eqref{approx2} is mathematically equivalent to maximizing a well-known submodular function, i.e., the facility location function \citep{cornuejols1977uncapacitated}.
\begin{definition}[Submodularity]\label{submod}
A set function $F: 2^{V} \rightarrow \mathbb{R}^{+}$ is submodular if $F(e \mid S) := F(S \cup\{e\})- F(S) \geq F(T \cup\{e\})-F(T)$, for any $S \subseteq T \subseteq V$ and $e \in V \backslash T$. $F$ is monotone if $F(e \mid S) \geq 0$ for any $e \in V \backslash \bar{S}$ and $S \subseteq V$. 
\end{definition}
According to Definition~\ref{submod} and eq.~\eqref{approx2}, we define a monotone submodular function $\mathcal{F}$ over $\mathcal{S}$ with estimated gradient $\Tilde{g}_i$:
$$\mathcal{F}(\mathcal{S}):= C -\sum_{j \in \mathcal{M}} \min _{i \in \mathcal{S}}\left\| \Tilde{g}_j -  \Tilde{g}_i\right\|$$
where $C$ is a constant to upper bound $\mathcal{F}(\mathcal{S})$. To formulate our task selection objective, we follow the logic of \citep{pooladzandi2022adaptive}. To restrict the size of $\mathcal{S}$, we limit the number of selected tasks in $\mathcal{S}$ to be no greater than $K$, i.e., $|\mathcal{S}| \leq K$. Thus, the submodular maximization form of selection objective is:
\begin{equation}
    \mathcal{S}^{\star} =\underset{\mathcal{S} \subseteq \mathcal{M}}{\arg \max } \mathcal{F}(\mathcal{S}), \text { s.t. }|\mathcal{S}| \leq K .
\end{equation}

The maximization problem of $\mathcal{F}(\mathcal{S})$ under cardinality constraint $|\mathcal{S}| \leq K$ has an approximation solution with $1-e^{-1}$ bound can be achieved via the greedy algorithm \citep{nemhauser1978analysis,wolsey1982analysis}. To start with, initialize $\mathcal{S}$ as an empty set $\emptyset$ and for each greedy iteration, merge an element $\mathcal{T} \in \mathcal{S}_{\mathcal{M}}^C$ that maximizes the marginal utility $\mathcal{F}(\mathcal{T}|\mathcal{S}) = \mathcal{F}(\mathcal{S} \cup \mathcal{T}) - \mathcal{F}(\mathcal{S})$. The update step for $\mathcal{S}$ can be described as $
    \mathcal{S} = \mathcal{S} \cup {\arg\max}_{\mathcal{T}_{j} \in \mathcal{S}_{\mathcal{M}}^C} \mathcal{F}(\mathcal{T}_{j} \mid \mathcal{S})$. The computational complexity of the entire greedy algorithm can be reduced to $\mathcal{O}(|\mathcal{M}|)$ using stochastic methods to choose random subset \citep{mirzasoleiman2015lazier}. This step is correspinds to line $6$ in \textbf{Algorithm \ref{alg:metatrain}}.

\subsection{DERTS with Noisy Tasks} \label{sub4}

Making the meta-learning model robust to label noise is an essential issue, as discussed in previous work \citep{mazumder2021rnnp,yao2021meta,liang2022few}. Previous work \citep{killamsetty2021glister,mirzasoleiman2020coresets2} claims that subset selection based on the gradient is robust to label noise in the standard noisy label setting. Specifically, the Jacobian matrix of a neural network can be well approximated by a low-rank matrix, suggesting a further claim that error for clean labels mainly lies in the subspace corresponding to the dominant  singular values while the error (corresponding to noisy labels) lies in the complementary subspace~\citep{mirzasoleiman2020coresets2}. Thus, the clean data potentially form clusters and aggregate with each other, while noisy labeled data spread out in the gradient space.

We propose a heuristic based on intuition: If a task has noisy label data, some of the labels are  incorrect, making the task more difficult to learn. As a result, the model may find it more difficult to find the correct parameters to minimize the loss, resulting in a larger gradient norm. On the other hand, if a task contains clean data, the labels are correct, and the model should be able to learn the task more efficiently, resulting in a smaller gradient norm. Based on the above motivation and the principle of not affecting the computation cost significantly, we dynamically set a threshold $h$ on the gradient norm to implicitly infer the task noise ratio by truncating tasks with a gradient norm higher than the set threshold. Suppose the truncated tasks set is $\mathcal{Z}:=\left\{\mathcal{T}_{l}|\Tilde{g}_{l} \geq h\right\}$, we take the complement of $\mathcal{Z}$ on $\mathcal{S}$ as the finalized subset for noisy task scenario. The algorithmic details can be found in \textbf{Algorithm \ref{alg:metatrain}}.

\subsection{Analysis for DERTS} \label{sub5}

We make three standard assumptions regarding the loss function $\mathcal{L}$: 

\begin{assumption}
\label{a1}
Every loss function $\mathcal{L}(f;\mathcal{D})$ is twice continuously differentiable with respect to $f$, and $\beta$-smooth, i.e.,
$$
\left\|\nabla \mathcal{L}(f;\mathcal{D})-\nabla \mathcal{L}(g;\mathcal{D})\right\| \leq \beta\|f-g\| .
$$
\end{assumption}
\begin{assumption}[\citep{karimi2016linear}]
\label{a2}
The loss function $\mathcal{L}(f;\mathcal{D})$ is $\mu-PL^{\star}$ (local PL condition), i.e.
$$
\frac{1}{2}\left\|\sum_{i \in \mathcal{M}}\nabla \mathcal{L}(f_i;\mathcal{D}_i)\right\|^2 \geq \mu \sum_{i\in\mathcal{M}}\mathcal{L}(f_i;\mathcal{D}_i), \quad \forall\mathcal{T}_{i} \in \mathcal{M}.
$$
\end{assumption}

\begin{assumption}
\label{a3}
The Hessian matrix of the loss function $\mathcal{L}(f;D)$ is bounded, i.e.
$$
m \leq \|\nabla^2 \mathcal{L}(f;\mathcal{D}) \| \leq M.
$$
\end{assumption}

\textbf{Assumption \ref{a1}} and \textbf{\ref{a2}} are common in the analysis of convergence of training dynamics \citep{oymak2019overparameterized,liu2022loss}. Due to the bilevel-optimization inherent to meta-learning, \textbf{Assumption \ref{a3}} is required -- a similar assumption is used in \cite{fallah2020convergence}. Our main analysis result concerning the error between loss function trained on the full task pools and the subsets is given below.

\begin{algorithm}[t!]
\small 
\caption{\small Efficient and Robust Task Selection (DERTS)}
\label{alg:metatrain}
\begin{algorithmic}[1]

\REQUIRE Initialized meta-model $f_{\theta_{0}}$; task distribution $p\mathcal{(T)}$; outer loop and inner loop learning rate $\eta_{1}, \eta_{2}$; number of selected tasks $K$; task pool $\mathcal{M}$; noisy setting flag $F$; noisy setting threshold $h$;
\STATE Initialize the meta-model $f_{\theta_{0}}$, task pool $\mathcal{M} \leftarrow \emptyset$, and task subset $\mathcal{S} \leftarrow \emptyset$
\FOR{Initialized Task Pool $\mathcal{M}$ :}
\STATE Randomly draw tasks $\mathcal{T}_i \sim p\mathcal{(T)}$ into task pool $\mathcal{M}$
\STATE Estimate outer loop query gradient $\Tilde{g}_{i}$ for every task $\mathcal{T}_i$ in the task pool $\mathcal{M}$
\WHILE{$|\mathcal{S}| < K$}
    \STATE $\mathcal{T}_{i} \in \arg \max _{\mathcal{T}_{i} \in \mathcal{S}_{\mathcal{M}}^C} \mathcal{F}(\mathcal{T}_{i} \mid \mathcal{S})$ 
    \STATE $\mathcal{S}=\mathcal{S} \cup\{\mathcal{T}_{i} \}$
    
\ENDWHILE
\STATE $\gamma_i=\sum_{j \in \mathcal{M}} \mathbbm{1}_{\left\{j=\operatorname{argmin}_{\mathcal{T}_{i} \in \mathcal{S}}\left\| \Tilde{g}_j - \Tilde{g}_i \right\|\right\}}$
\IF{noisy setting flag $F$}
    \STATE Initialize drop task set $\mathcal{Z} \leftarrow \emptyset$
    \FOR{$\mathcal{T}_{j} \in \mathcal{S}$}

    \IF{$\Tilde{g}_{j} \geq h$}
    \STATE $\mathcal{Z}=\mathcal{Z} \cup\{\mathcal{T}_{j} \}$
    \ENDIF
    \ENDFOR
\STATE $\mathcal{S} = \mathcal{Z}_{\mathcal{S}}^C$
\ENDIF

\STATE \texttt{meta-learn}($\mathcal S, \gamma_1,\hdots, \gamma_{|K|}$) Run standard meta-learning algorithm on $\mathcal{S}$ with weights $\{\gamma_{1},\cdots,\gamma_{|K|}\}$ multiply the corresponding computed outer loop gradient of each task.
\ENDFOR

\end{algorithmic}

\end{algorithm}

\begin{theorem}[Training Dynamics]
\label{Theorem 1}
Assume that the loss function $\mathcal{L}(f,\mathcal{D})$ satisfies  assumptions \ref{a1} --\ref{a3} and $\epsilon$ is an upper bound for the RHS of Eq.\eqref{approx}. Then, with the proper constant learning rate $\eta$ and $\eta'$ for outer and inner loop updates and a initialization point $\theta^0$, applying DERTS  has similar training dynamics to that of training on the full task pool $\mathcal{M}$. Specifically,
\begin{equation}\label{theo}
\begin{split}
    \mathbb{E}[\left(\mathcal{L}(f_{\theta^{t}};\mathcal{D})\right)]\ 
    \leq (1 -\eta\eta'm\mu)^t \mathbb{E}[(\mathcal{L}(f_{\theta^{0}};\mathcal{D}))] \\
    + \frac{1}{\eta\eta'm\mu}(\eta \epsilon \|\mathbb{E}[\nabla_{\theta} \mathcal{L}(f_{\phi};\mathcal{D})]\|_{L^{\infty}_\mathcal{M}} - \frac{\eta}{2} \epsilon^2) +  r 
\end{split}
\end{equation}
where 
\begin{equation*}
\begin{split}
    &\|\mathbb{E}[\nabla_{\theta}  \mathcal{L}(f_{\phi};\mathcal{D})]\|_{L^{\infty}_\mathcal{M}} =\\ 
    & \sup_{\Gamma}\left\|\mathbb{E}\left[\sum_{j \in \mathcal{M}} \left( \nabla_{\theta}\mathcal{L}\left(f_{\phi_{j}} ; \mathcal{\mathcal{D}}_{j}^q\right)  -\nabla_{\theta}\mathcal{L}\left(f_{\phi_{\Gamma(j)}} ; \mathcal{\mathcal{D}}_{\Gamma(j)}^q \right)\right)\right]\right\|,
\end{split}
\end{equation*}
and 
$$
r = |\mathcal{L}(f_{\phi*},\mathcal{D}) - \mathcal{L}(f_{\theta^0},\mathcal{D})|.
$$

\end{theorem}

This result implies training dynamics between subset learning and task pool learning is bounded. The first term on the RHS of~\eqref{theo} is a contraction that decreases with each iteration. The second term shows the difference between training on subsets $\mathcal{S}$ and all the tasks. The last term $r$ is a bias  resulting from the bilevel optimization for meta-learning. We note that  the bias  is not an artefact of the DERTS algorithm but is inherent in the meta-learning formulation. We show the proof in \textbf{Appendix C}.

\section{Experiment Results}

\begin{table*}[t]

\centering
\setlength{\tabcolsep}{4pt}
\begin{tabular}{l|lll|lll}
\hline \hline
Dataset & \multicolumn{3}{l|}{Mini-ImageNet (5-way 5-shot)} & \multicolumn{3}{l}{Mini-ImageNet (5-way 1-shot)}\\
\hline
Iterations         & 1000  & 3000 & 10000 (All)  & 500 & 1500 & 5000 (All) \\ \hline
ANIL     &  $56.50 \pm 0.8$    &  $61.39 \pm 0.6$    &   $62.83 \pm 0.6$      &   $41.04 \pm 0.8$  &   $46.06 \pm 0.7$    &   $47.22 \pm 0.7$         \\ 
ANIL-US        &   $54.32 \pm 0.8$   &   $58.92 \pm 0.7$   &    $62.04 \pm 0.6$      &   $37.01 \pm 0.8$   &   $43.58 \pm 0.8$   &  $45.94 \pm 0.8$             \\ 
ANIL-ATS        &  $52.97 \pm 0.9$    &  $60.74 \pm 0.7$    &    $61.82 \pm 0.8$      &  $42.38 \pm 0.7$    &   $46.62 \pm 0.6 $   &  $46.74 \pm 0.7$         \\ 
\textbf{ANIL-DERTS}      &   $\mathbf{57.67 \pm 0.6}$    &  $\mathbf{61.78 \pm 0.8}$    &  $\mathbf{63.07 \pm 0.7}$  &    $\mathbf{44.06 \pm 0.8}$  &  $\mathbf{47.07 \pm 0.6}$   &  $\mathbf{47.67 \pm 0.7}$  \\ \hline
ProtoNet(PN) &  $60.19 \pm 0.8$    &  $62.41 \pm 0.7$    &   $\mathbf{66.29 \pm 0.8}$   &  $41.97 \pm 0.8$ &  $43.97 \pm 0.6$    &  $48.32 \pm 0.7$  \\
PN-US    &  $57.21 \pm 1.0$    &  $61.79 \pm 0.9$    &    $64.50 \pm 0.9$  &  $40.90 \pm 0.6$  &   $44.00 \pm 0.8$   &  $46.50\pm 0.8$  \\ 
\textbf{PN-DERTS}   &  $\mathbf{62.16 \pm 0.6}$    &  $\mathbf{65.09 \pm 0.7}$    &   $65.51 \pm 0.7$  &   $\mathbf{46.18 \pm 0.6}$   &   $\mathbf{47.08 \pm 0.8}$   &  $\mathbf{48.52 \pm 0.6}$\\ \hline

\hline
Dataset & \multicolumn{3}{l|}{Tiered-ImageNet (5-way 5-shot)} & \multicolumn{3}{l}{Tiered-ImageNet (5-way 1-shot)}\\
\hline
Iterations          & 1000& 3000 & 10000 (All)  & 500 & 1500 & 5000 (All)  \\ \hline
ANIL     &  $56.92 \pm 0.8$    &  $62.48 \pm 0.6$    &   $63.70 \pm 0.7$     &   $40.59 \pm 0.6$   &  $46.03 \pm 0.7$  &       $48.28 \pm 0.7$     \\ 
ANIL-US        &  $56.06 \pm 1.0$    &   $58.43 \pm 0.8$   &   $62.20 \pm 0.8$       &  $36.87 \pm 0.8$    &  $43.08 \pm 0.9$    &      $46.15 \pm 0.8$      \\ 
ANIL-ATS        &  $53.65 \pm 0.9$    &   $62.92 \pm 0.8$   &    $62.34 \pm 0.7$      &   $42.74 \pm 0.7$   &  $46.55 \pm 0.7$    &     $47.81 \pm 0.8$      \\ 
\textbf{ANIL-DERTS  }    &   $\mathbf{60.49 \pm 0.7}$    &  $\mathbf{63.07 \pm 0.8}$    &  $\mathbf{64.01 \pm 0.8}$  &     $\mathbf{44.47 \pm 0.8}$   &    $\mathbf{47.78 \pm 0.8}$   &  $\mathbf{48.40 \pm 0.8}$   \\ \hline
ProtoNet(PN) &   $64.28 \pm 0.7$   &    $65.22 \pm 0.8$  &   $\mathbf{69.26\pm0.8}$   &  $43.52 \pm 0.7$  &  $47.07 \pm 0.7$    &  $\mathbf{49.56 \pm 0.7}$  \\
PN-US    &  $60.67\pm0.8$    &  $64.05 \pm 0.8$    &   $67.89\pm 0.8$   &  $40.43 \pm 0.7$  &  $42.73 \pm 0.8$    &  $49.02\pm0.8$   \\ 
\textbf{PN-DERTS}   &   $\mathbf{65.02 \pm 0.7}$   &    $\mathbf{66.03 \pm 0.8}$  &  $69.11 \pm 0.7$   &   $\mathbf{44.19 \pm 0.7}$   &  $\mathbf{47.96 \pm 0.7}$    &  $49.48 \pm 0.7$\\ \hline \hline
\end{tabular}
 
\caption{ Average accuracy (\%) of 5-way 5-shot and 5-way 1-shot Mini-ImageNet and Tiered-ImageNet Classification with Limited Budget Setting. In the 5-way 5-shot setting, iterations of 1000 (3000) in the table denote the performance after learning on 10\% (30\%) tasks during the episodic training. In the 5-way 1-shot setting, iterations of 500 (1500) in the table denote the performance after learning on 10\% (30\%) tasks during the episodic training. } 
\label{tiered}
\vspace{-0.3cm}
\end{table*}

Here we demonstrate the effectiveness of DERTS by performing extensive experiments on widely used image classification benchmarks in two settings: (i) limited data budget and (ii) noisy label tasks. 

\vspace{-0.2cm}


\paragraph{Dataset and Baseline:} We conduct experiments on the standard image classification benchmarks: \textbf{Mini-ImageNet}\citep{vinyals2016matching} and \textbf{Tiered-ImageNet}\citep{ren2018meta}. The tasks are constructed as $5$-way $1$-shot and $5$-way $5$-shot for limited data budget setting and $5$-way $5$-shot for noisy task setting. To show the generality of DERTS, we implemented the proposed algorithm on both gradient-based meta-learning method ANIL \citep{raghu2020rapid} and metric-based meta-learning method ProtoNet (PN) \citep{snell2017prototypical}. For sampling strategies, we compare DERTS with the state-of-the-art sampling paradigm for meta-learning Uniform Episodic Sampling (US) \citep{arnold2021uniform} and Adaptive Task Scheduler (ATS) \citep{yao2021meta}. According to the sampling mechanism, we combine US with both ANIL and PN and ATS for ANIL (ATS is designed only for gradient-based meta-learning algorithms).

\vspace{-0.2cm}

\paragraph{Implementation Details:} For both ANIL and PN,  We use a 4-layer Convolutional Neural Network (CNN4) as the model backbone. Additional experiments with ResNet-12 \citep{oreshkin2018tadam} could be found in \textbf{Appendix E}. We set the meta-batch size to 32.   
Adam \citep{kingma2014adam} is selected as the optimizer. We set the learning rates for ANIL as $0.005 - 0.01$ for outer loops and $0.5$ for inner loops with $3$ adaptation steps. For PN, we set the learning rate as $0.005$. We fix the episodic task pools with the size of $3200$, and the number of selected tasks for each pool is $960$ (meta-training on each selected subset would take $30$ iterations based on a batch size of $32$). All experiments are conducted on NVIDIA A6000 GPU.

\vspace{-0.2cm}

\subsection{Meta-Learning with Limited Budget}

\vspace{-0.2cm}

\paragraph{Experiment Setup:} In this setting, we consider the following questions: (i) Assuming no modification on task generation, can DERTS achieve comparable performance with fewer tasks? (ii) Under the scenario where we dramatically reduce the number of meta-training classes (i.e., 64 classes reduced to 16 classes for the training set of Mini-ImageNet), we then ask how DERTS performs.

We train the meta-model for 10000 iterations for $5$-way $5$-shot setting and 5000 iterations for $5$-way $1$-shot setting. We set a $500$ iteration warm-up process for ANIL and a $100$ iteration warm-up process for PN.

\begin{table}

\centering
\renewcommand{\arraystretch}{0.8}
\setlength{\tabcolsep}{4pt}
\begin{tabular}{l|l|l|l|l}
\hline \hline
        Time              & ANIL & ANIL-US & ANIL-ATS & \textbf{ANIL-DERTS} \\ \hline
[s] &       0.60      &  0.69        &    3.11   &    0.95  \\ \hline \hline
\end{tabular}
\caption{ Average single iteration run time for limited budget setting.}
\label{time}
\end{table}

\paragraph{Experiment Results with Fewer Tasks:} In \textbf{Table \ref{tiered}} we show the average accuracy with $95\%$ confidence interval of training a CNN4 base-model on Mini-ImageNet and Tiered-ImageNet for ANIL and PN after varying number of iterations. We evaluate checkpoints after training $10\%$ of all iterations (1000 for 5-way 5-shot and 500 for 5-way 1-shot) and $30\%$ of all iterations (3000 for 5-way 5-shot and 1500 for 5-way 1-shot). Our key observations are as follows. (i) \textbf{DERTS is data-efficient for episodic training}. For all the evaluation after $10\%$ and $30\%$ of all iterations for both 5-way 5-shot and 5-way 1-shot scenarios, DERTS achieves higher performance than US, ATS, and vanilla ANIL and PN with random sampling by average gain $4.52\%$ against US, $2.23\%$ against ATS, and $1.92\%$ against vanilla ANIL and PN on accuracy. 6 out of 8 of the evaluation after all iterations, DERTS outperforms all the baselines, and the rest 2 evaluation is averagely 0.46\% behind the best performer. The fact that DERTS's full iteration performance indicates the diversity of selected tasks, preserves the generalization capacity for the meta-model. (ii)\textbf{DERTS gains significant speedup against other sampling strategies}. \textbf{Table \ref{time}} shows the average per iteration running time for ANIL-based algorithm on Mini-ImageNet. ANIL has the fastest running time because no extra module is added to the vanilla algorithm. Although US is the fastest sampling strategy among the three modified methods, from \textbf{Table \ref{tiered}} we can tell that US has the lowest convergence and performance on accuracy. As ATS has a significantly heavy computation load, the running time of ATS is above 4 times of DERTS's. The execution time of DERTS is slightly higher than vanilla ANIL and US, suggesting that DERTS's estimation of task gradient and submodular maximization with the stochastic greedy algorithm are efficient. Combining the results from \textbf{Table \ref{tiered}} and \textbf{Table \ref{time}}, we claim that DERTS gains general speedup on computation time against baselines for reaching a comparable performance.

\begin{table}
\vspace{-0.2cm}
\centering

\setlength{\tabcolsep}{4pt}
\begin{tabular}{l|ll}
\hline \hline
Dataset & \multicolumn{2}{l}{Mini-ImageNet (16-Class)} \\
\hline
Method          & 5-way 5-shot & 5-way 1-shot\\ \hline
ANIL     &  $45.97 \pm 0.65$    &  $33.61 \pm 0.66$    \\ 
ANIL-US        &   $46.70 \pm 0.70$   &   $34.70 \pm 0.50$         \\ 
ANIL-ATS        &  $47.76 \pm 0.68$    &  $35.15 \pm 0.67$       \\ 
\textbf{ANIL-DERTS}      &   $\mathbf{48.92 \pm 0.54}$    &  $\mathbf{36.85 \pm 0.75}$   \\ \hline 
ProtoNet        &   $50.27 \pm 0.67$   &   $36.46 \pm 0.67$         \\ 
ProtoNet-US        &  $49.21 \pm 0.77$    &  $35.48 \pm 0.70$       \\ 
\textbf{ProtoNet-DERTS}      &   $\mathbf{52.08 \pm 0.71}$    &  $\mathbf{37.59\pm 0.62}$   \\ \hline \hline
\end{tabular}
\caption{ 5-way 5-shot / 1-shot Mini-ImageNet Classification with 25\% Class Training Set.}
\label{25}
\vspace{-0.3cm}
\end{table}


\paragraph{Experiment Results with Fewer Class for Generating Tasks:} We conduct experiments on the setting of a limited budget of training classes proposed by ATS \citep{yao2021meta}. In the few-shot classification problem, each training episode is a few-shot task that subsamples classes as well as data points. In mini-Imagenet, the original number of meta-training classes is 64, corresponding to more than 7 million 5-way combinations. Thus, we control the budgets by reducing the number of meta-training classes to 16, resulting in 4,368 combinations. Thus, in this new setting, not only did the data for constructing tasks decrease to 25\% in the original setting, but also the number of classes decreased to 25\% of the original setting, imposing a significant decrease in task diversity.

\textbf{Table \ref{25}} shows the result of the aforementioned setting with fewer classes. By only training on 25\% of the classes, we observe that DERTS outperforms all the baselines with an average of $2\%$ on accuracy. Due to the significant decrease in task diversity, we claim our task selection method still correctly captures the decreased diversity and focuses on the more informative tasks for learning representation. Additional experiments with other ratios of decreased meta-training class can be found in \textbf{Appendix E}.

\subsection{Meta-Learning with Noisy Tasks}

\begin{table*}[h]

\centering
\setlength{\tabcolsep}{4pt}
\begin{tabular}{l|ll|ll}
\hline
\hline
Dataset & \multicolumn{2}{l|}{Mini-ImageNet (5-way 5-shot)} & \multicolumn{2}{l}{Tiered-ImageNet (5-way 5-shot)}\\
\hline
Noise Ratio          & 25\% & 40\% & 25\% & 40\% \\ \hline
ANIL     &    $61.12 \pm 0.85$    &  $57.32 \pm 0.79$       &     $60.84 \pm 0.65$    &      $57.31 \pm 0.78$      \\ 
ANIL-US        &   $58.48 \pm 0.69$   &    $52.39 \pm 0.90$      &   $57.13 \pm 0.75$   &     $51.11 \pm 1.21$       \\ 
ANIL-ATS        &   $59.46 \pm 0.72$     &    $54.42 \pm 0.73$      &     $58.35 \pm 0.79$     &     $54.73 \pm 0.72$      \\ 
\textbf{ANIL-DERTS}      &    $\mathbf{62.38 \pm 0.66}$   &  $\mathbf{59.14 \pm 0.72}$       &  $\mathbf{62.40\pm 0.70}$    &  $\mathbf{58.23 \pm 0.81}$ \\ \hline 
ProtoNet &      $58.68 \pm 0.85$   &   $49.85 \pm 0.85$    &   $59.34 \pm 0.79$   &  $50.02 \pm 0.91$  \\ 
ProtoNet-US    &  $56.09 \pm 0.82$    &  $46.96 \pm 0.80$    &    $55.16 \pm 0.76$  &    $45.99 \pm 0.98$      \\

\textbf{ProtoNet-DERTS}   &   $\mathbf{60.98 \pm 0.84}$   &  $\mathbf{51.40 \pm 0.78}$    &  $\mathbf{60.77 \pm 0.81}$    &  $\mathbf{52.58 \pm 0.79}$    \\ \hline\hline
\end{tabular}
\caption{ 5-way 5-shot Mini-ImageNet and Tiered-ImageNet Classification with Noisy Tasks. 25\% and 40\% denote the noise ratio (percentage of mislabeled data).}
\vspace{-0.3cm}
\label{noise1}
\end{table*}

\paragraph{Experiment Setup:} Unlike previous work \citep{yao2021meta,liang2022few} which only constructs noisy data within the support set, we extend the noisy data to both the support set and query set, without providing the noisy data's specific identity. The details of generating noisy tasks can be found in \textbf{Appendix D}. We consider two specific setups: the first where 25\% of data points are incorrectly labeled, the second where 40\% are mislabelled (25\% setting means the average noise ratio for all the tasks is 25\%, but there could be extremely high noise ratio tasks and roughly clean-label tasks). We test 5-way 5-shot setting with symmetric label flipping \citep{van2015learning}. For the $25\%$ case, we use the same warm-up model as with the limited budget setting. For the $40\%$ case, we double the number of warm-up iterations for ANIL and PN. We set the estimated gradient norm threshold $h$ as 1.25 times the average of the task estimated gradient norm in the task pools.

\begin{figure}
\vspace{-0.5cm}
\centering
\begin{subfigure}[t]{0.42\linewidth}
\centering
\includegraphics[width=3.9cm]{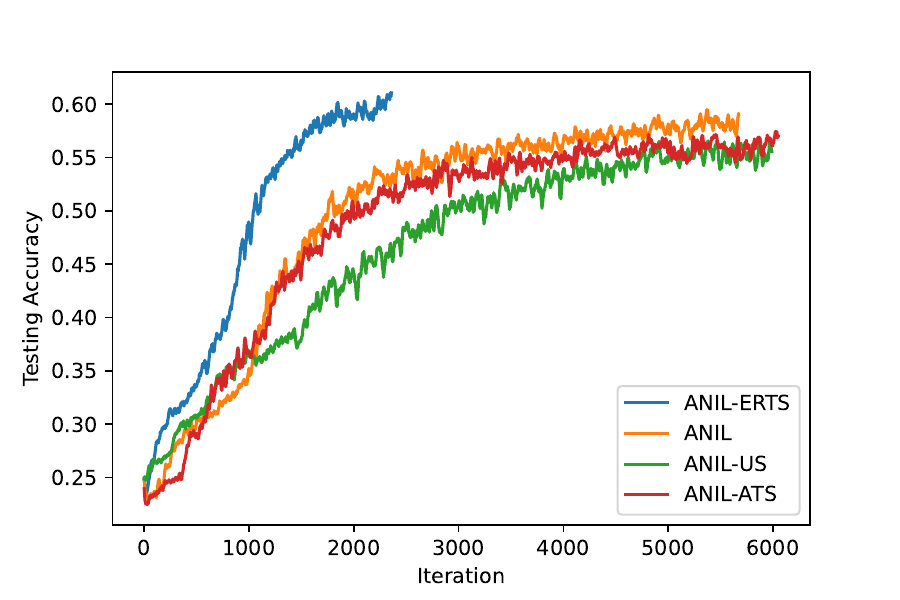}
\caption{}
\label{fig:domain-aware-expand-a}
\end{subfigure}
\begin{subfigure}[t]{0.42\linewidth}
\centering
\includegraphics[width=3.9cm]{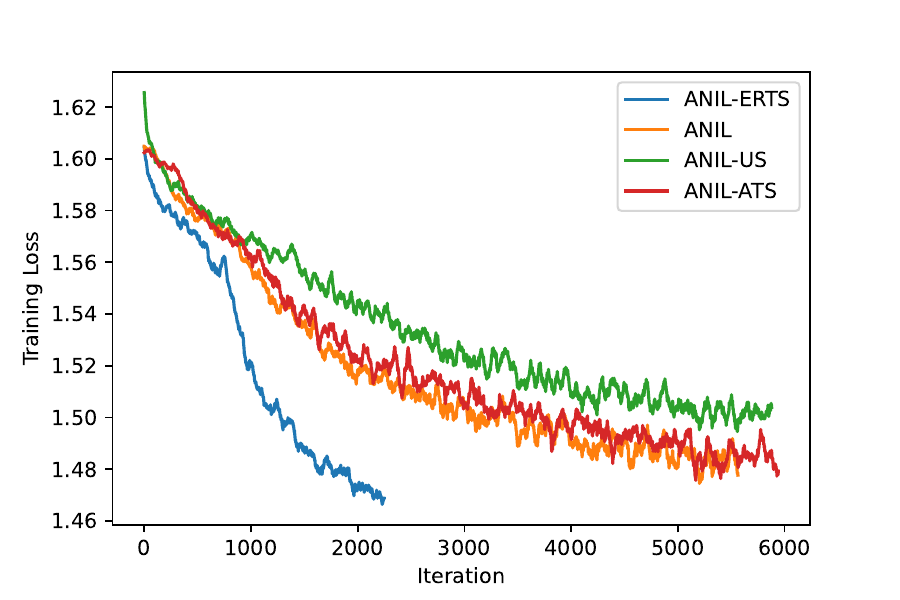}
\caption{}
\label{fig:domain-aware-expand-b}
\end{subfigure}

\begin{subfigure}[t]{0.42\linewidth}
\centering
\includegraphics[width=3.9cm]{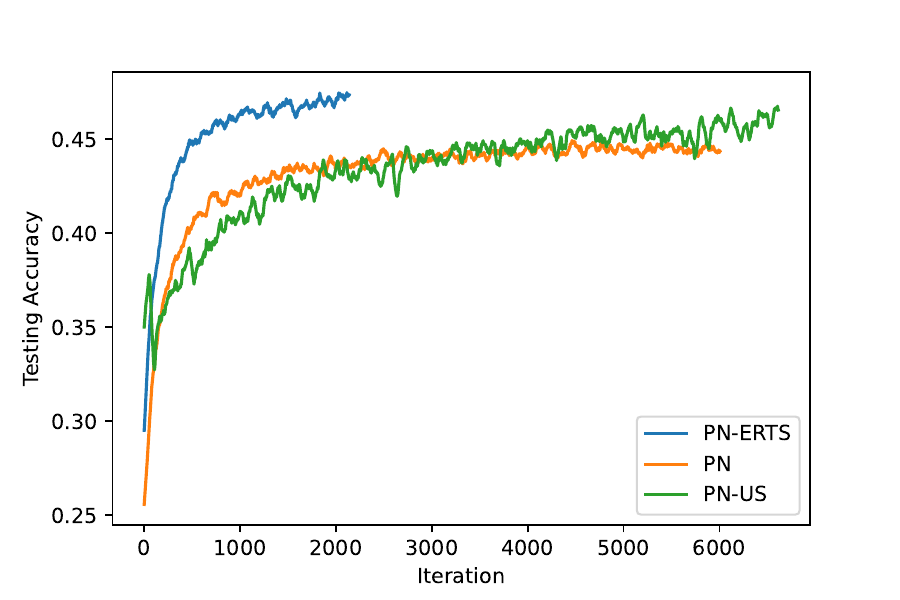}
\caption{}
\label{fig:domain-aware-expand-c}
\end{subfigure}
\begin{subfigure}[t]{0.42\linewidth}
\centering
\includegraphics[width=3.9cm]{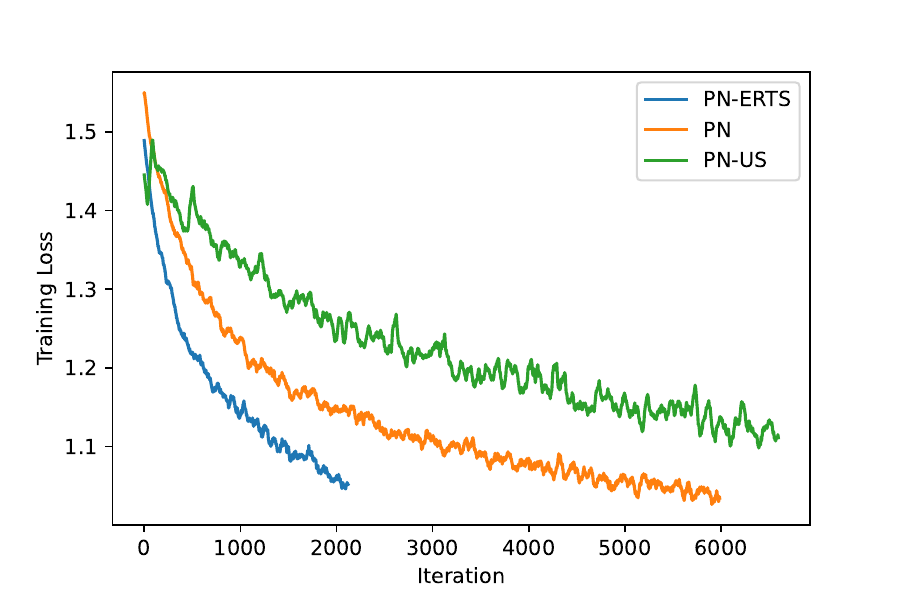}
\caption{}
\label{fig:expandnet}
\end{subfigure}
\vspace{-0.2cm}
\caption{\small Loss Residual and Accuracy for Noisy Task Settings (Early Stage). (a) Test Accuracy of $25\%$ Noise Setting on Mini-ImageNet of ANIL. (b) Training Loss of $25\%$ Noise Setting on Mini-ImageNet of ANIL. (c) Test Accuracy of $40\%$ Noise Setting on Mini-ImageNet of PN. (d) Training Loss of $40\%$ Noise Setting on Mini-ImageNet of PN.}
\label{sp}
\vspace{-0.2cm}
\end{figure}

\vspace{-0.3cm}
\paragraph{Experiment Result:} \textbf{Table \ref{noise1}} shows the average accuracy with a $95\%$ confidence interval of training a CNN4 base-model on Mini-ImageNet and Tiered-ImageNet for ANIL and ProtoNet with 12000 iterations. DERTS outperforms all baselines in the noisy task settings, by achieving on average $3\%$ improvement in testing accuracy. It is worth noting that in the $40\%$ settings, DERTS acquires a larger gain on performance than in $25\%$ noisy setting, which empirically demonstrates the robustness of our algorithm. Another important observation is that ATS and US are empirically sensitive to the noisy task setting. The reason  ATS is less robust than vanilla ANIL and ANIL-DERTS may be caused by the inner product computation for gradient on support set and query set is ineffective because both support sets and query sets contain the noisy labeled data. \textbf{Figure \ref{sp}} shows the training dynamics of DERTS and baselines. We observe that aside from the final gain in performance, DERTS converges faster than all baselines as well.

\subsection{Ablation Study on DERTS}

\begin{table}

\centering

\renewcommand{\arraystretch}{0.8}
\setlength{\tabcolsep}{4pt}
\begin{tabular}{l|l}
\hline \hline
Selecting Ratio $k/N$                       & Accuracy \\ \hline
10\%                            &   $61.70 \pm 0.72$       \\ 
20\%                            &   $62.07 \pm 0.77$       \\ 
30\% (Default)                  &  $\mathbf{63.07 \pm 0.71}$        \\ 
30\% (No Weights) &  $59.55 \pm 0.72$        \\ 
50\%                            &    $62.98 \pm 0.70$      \\ \hline \hline
\end{tabular}
\caption{ Ablation Study on Selection Ratio and Weights with Mini-ImageNet 5-way 5-shot Classification}
\vspace{-0.3cm}
\label{ablation}
\end{table}

We investigate the effectiveness of selecting ratio $k/N$ ($k$ is the size of subsets and $N$ is the size of task pools) and weights $\left\{\gamma_{i} ~|~ i = 1,2,\ldots,k\right\}$ in the ablation study.

\textbf{Table \ref{ablation}} shows the results of different selecting ratio $k/N$ and the default ratio without adding weights. We observe that with a low selecting ratio, DERTS's performance drops 1\%-2\%. This phenomenon may caused by the decrease of selected tasks' diversity since the selected ratio $k/N$ is low and the subset can not include all the informative tasks from the task pool. The performance of setting without adding weights is significantly lower than the setting with corresponding weights, showing the selected tasks are not equally important as well. This phenomenon suggests DERTS's weight selection is an essential element for capturing the full training dynamics. Additional ablation study can be found in \textbf{Appendix E}.

\section{Conclusion \& Discussion}

We propose Data-Efficient and Robust Task Selection (DERTS), a task selection algorithm for meta-learning from the view of optimization to address the issue of data efficiency, and robustness against noisy labels. By selecting subsets with weights to approximate the gradient of the tasks from the task pools, DERTS is data-efficient for episodic training under the limited budget scenario. We then modify DERTS to address the robustness of noisy data scenarios. Our extensive experiments demonstrate that DERTS outperforms the state-of-the-art sampling and selection strategies in terms of both data efficiency and robustness. 

\section{Acknowledgements} JA acknowledges funding from NSF grants ECCS 2144634 and ECCS 2047213 as well as the Columbia Data Science Institute.

{
    \small
    \bibliographystyle{ieeenat_fullname}
    \bibliography{main}
}

\newpage
\onecolumn 

\vspace{0.5in}
\begin{center}
 \rule{6.875in}{0.7pt}\\ 
 {\Large\bf Supplementary Material for\\ ``Data-Efficient and Robust Task Selection for Meta-Learning ''}
 \rule{6.875in}{0.7pt}
\end{center}
\appendix

\section{Introduction to Meta-Learning Algorithms} \label{meta-intro}

\paragraph{Gradient-based meta-learning}
The goal of gradient-based meta-learning is to learn initial parameters $\theta^*$ such that taking one (or a few) gradient steps on the support set $\mathcal{D}^s$ leads to a model that performs well on task $\mathcal{T}$.  Consider model-agnostic meta-learning (MAML) \citep{finn2017model} with base model $f$ as an illustrative example.  During the meta-training stage, the performance of the adapted model $f_\phi$ (i.e., $ \phi=\theta-\eta \nabla_\theta \mathcal{L}\left(f_\theta ; \mathcal{D}^s\right)$, $\eta$ denotes the inner-loop learning rate) is evaluated on the corresponding query set $\mathcal{D}^q$ and is used to optimize the model parameter $\theta$. Formally, the bi-level optimization process with expected risk is formulated as: 
\begin{equation*}
\theta^* \leftarrow \arg \min _\theta \mathbb{E}_{\mathcal{T} \sim p(\mathcal{T})}\left[\mathcal{L}\left(f_\phi; \mathcal{D}^q\right)\right].
\end{equation*}

During the meta-testing stage, for task $\mathcal{T}_t$, the adapted parameter $\phi_t$ is found by fine-tuning meta-model $\theta_t$ on the support set $\mathcal{D}_t^s$. The almost no inner loop (ANIL) algorithm \citep{raghu2020rapid} simplifies the inner loop computation by only updating the classification head during meta-training task adaptation while keeping the remainder frozen. ANIL achieves a comparable performance with MAML with lower computational cost.

\paragraph{Metric-based meta-learning}
The aim of metric-based meta-learning is to conduct a non-parametric learner on top of meta-learned embedding space. Taking prototypical network (ProtoNet) with base model $f_{\theta}$ as an example \citep{snell2017prototypical}, for each task $\mathcal{T}$, we first compute a class prototype representation $\left\{\mathbf{c}_r\right\}_{r=1}^R$ as the representation vector of the support samples belonging to class $k$ as $\mathbf{c}_r=\frac{1}{N_r} \sum_{\left(\mathbf{x}_{k ; r}^s, \mathbf{y}_{k ; r}^s\right) \in \mathcal{D}_r^s} f_\theta^{P N}\left(\mathbf{x}_{k ; r}^s\right)$, where $\mathcal{D}_r^s$ represents the subset of support samples labeled as class $r$, $\left(\mathbf{x}_{k ; r}^s, \mathbf{y}_{k ; r}^s\right)$ denotes the data with corresponding label in $\mathcal{D}_r^s$, and the size of this subset is $N_r$. Then, given a query data sample $\mathbf{x}_k^q$ in the query set, the probability of assigning it to the $r$-th class is measured by the distance $d$ between its representation $f_\theta\left(\mathbf{x}_k^q\right)$ and prototype representation $\mathbf{c}_r$, and the cross-entropy loss of ProtoNet is formulated as:
$$
\begin{aligned}
    \mathcal{L} =\mathbb{E}_{\mathcal{T} \sim p(\mathcal{T})}\left[-\sum_{k, r} \log \frac{\exp \left(-d\left(f_\theta\left(\mathbf{x}_k^q\right), \mathbf{c}_r\right)\right)}{\sum_{r^{\prime}} \exp \left(-d\left(f_\theta\left(\mathbf{x}_k^q\right), \mathbf{c}_{r^{\prime}}\right)\right)}\right]
\end{aligned}
$$
At the meta-testing stage, the predicted label of each query sample is assigned to the class with maximal probability, i.e., $\hat{\mathbf{y}}_k^q=\arg \max _r p\left(\mathbf{y}_k^q=r \mid \mathbf{x}_k^q\right)$.

\section{Efficient Gradient Estimation}\label{propproof}

The updating process of the meta-model with explicit task gradient $\nabla_{\theta}\mathcal{L}\left(f_{\phi_{j}}; \mathcal{D}_{j}^q\right)$ is time-consuming and incurs a large computational cost. As shown in \cite{katharopoulos2018not}, the variation of the gradient norm is mainly captured by the gradient of the loss function with respect to the pre-activation outputs of the last layer. Therefore, for a few-shot classification task, the above estimation only requires a forward computation on the last layer. E.g., for a softmax layer as the last, the gradients of the loss with respect to the input of the softmax layer for $\left(\mathbf{x}_{i, j}^q, \mathbf{y}_{i, j}^q\right)$ is $l_{i}$-$y_{i}$, where $l_{i}$ is the logits and $y_{i}$ is the encoded label. In this section, we elaborate on the details of our efficient gradient estimation as mentioned in sec. 4.2. We extend the result of \cite{pooladzandi2022adaptive} to estimate the task-gradient $\nabla_{\theta}\mathcal{L}\left(f_{\phi_{i}}; \mathcal{D}_{i}^q\right)$ and denote it as $\Tilde{g}_i$.

Generally, we follow the notation of \citep{katharopoulos2018not,pooladzandi2022adaptive} to establish our analysis upon the estimated gradient. Suppose in a $L$-layer multilayer perceptron network, $\theta^{(l)} \in \mathbb{R}^{M_l \times M_{l-1}}$ denotes the weight matrix for layer $l$  with $M_{l}$ hidden units and $\sigma^{(l)}(\cdot)$ be a Lipschitz continuous activation function. Then, for datapoint $\left(x_{i},y_{i} \right)$, let
$$
\begin{aligned}
x_{i}^{(0)} & =x_{i} \\
z_{i}^{(l)} & =\theta^{(l)} x_{i}^{(l-1)} \\
x_{i}^{(l)} & =\sigma^{(l)}\left(z_{i}^{(l)}\right) \\
f_{\theta}(x) & =x^{(L)}
\end{aligned}
$$
where $x_{i}^{(l)}$ denotes the output after the $l$-th layer of $x_{i}$ ($l=1,\hdots, L$). 
The gradient of the loss w.r.t. the output of the network is shown to be
$$
\nabla^{(i)}_{\theta^{(L)}} \mathcal{L}=\nabla_{\theta^{(L)}} \mathcal{L}\left(f_{\theta}\left(x_i \right), y_i\right)
$$
and the gradient of the loss w.r.t. the output of layer $l$ is 
$
\nabla^{(i)}_{\theta^{(l)}} \mathcal{L}=\Delta_i^{(l)} \Sigma_L^{\prime}\left(z_i^{(L)}\right) \nabla^{(i)}_{\theta^{(L)}} \mathcal{L}
$
where

\begin{equation*}
\Delta_i^{(l)}=\Sigma_l^{\prime}\left(z_i^{(l)}\right) \theta_{l+1}^T \ldots \Sigma_{L-1}^{\prime}\left(z_i^{(L-1)}\right) \theta_L^T
\quad \text{and} \quad
\Sigma_l^{\prime}(z)=\operatorname{diag}\left(\sigma^{\prime(l)}\left(z_1\right), \ldots, \sigma^{\prime(l)}\left(z_{M_l}\right)\right).
\end{equation*}

Thus, datapoint $i$'s gradient w.r.t. the parameters of the
$l$-th layer $\theta^{(l)}$ can be written as
$$
\begin{aligned}
\nabla_{\theta^{(l)}} \mathcal{L}\left(f_{\theta}\left(x_i \right), y_i\right) =\left(\Delta_i^{(l)} \Sigma_L^{\prime}\left(z_i^{(L)}\right) \nabla^{(i)}_{\theta^{(L)}} \mathcal{L}\right)\left(x_i^{(l-1)}\right)^T
\end{aligned}.
$$

For a query set $\mathcal{D}^{q}_{i}$ of arbitrary task $\mathcal{T}_{i}$, the gradient of meta-model $f_{\theta} $ on $\mathcal{D}^{q}_{i}$ w.r.t.  the $l$-th layer $\theta^{(l)}$ is
$$
\nabla_{\theta^{(l)}} \mathcal{L}(f_{\theta}; \mathcal{D}_i^q) = \sum_{k} 
\nabla_{\theta^{(l)}} \mathcal{L}\left(f_{\theta}\left(x_{ik} \right), y_{ik}\right)
$$
where $x_{ik}$ and $y_{ik}$ are the data and corresponding labels within query set $\mathcal{D}^{q}_{i}$.

Specifically, following previous work, we use the below gradient estimation $\Tilde{g}_i$ to approximate the gradient: $$\Tilde{g}_{i} = \sum_{k}\Sigma_L^{\prime}\left(z_{ik}^{(L)}\right) \nabla^{(ik)}_{\theta^{(L)}} \mathcal{L},$$
where $L$ denotes the last layer. 

Next, in \textbf{Proposition \ref{gradbound}} we show how to efficiently bind the task gradient with the adapted model via the gradient of loss w.r.t. the input of the last layer. 

\begin{proposition}[Gradient Norm Upper Bound]
Suppose the loss function $\mathcal{}$ is $\beta$-smooth, the norm of difference of task-specific meta-gradients $\nabla_{\theta}\mathcal{L}\left(f_{\phi_{i}}; \mathcal{D}_{i}^q\right)$ and $\nabla_{\theta}\mathcal{L}\left(f_{\phi_{j}}; \mathcal{D}_{j}^q\right)$ can be upper bounded by a constant $C_{1}$ times the norm of difference of $\Tilde{g_{i}}$ and $\Tilde{g_{j}}$ (gradients of the last layer of meta-model $\theta$) with adding another constant $C_{2}$, i.e.,
\begin{equation*}
\left\|\nabla_{\theta} \mathcal{L}(f_{\phi_{i}};  \mathcal{D}_i^q) - \nabla_{\theta} \mathcal{L}(f_{\phi_{j}}; \mathcal{D}_j^q)\right\| \leq C_{1} \left\| \Tilde{g}_{i} - \Tilde{g}_{j}  \right\| + C_{2}.
\end{equation*}
\label{gradbound}
\end{proposition}

Consider query sets $\mathcal{D}^{q}_{i}$ (with the same size) and $\mathcal{D}^{q}_{j}$ for two different tasks $\mathcal{T}_{i}$ and $\mathcal{T}_{j}$, we have 
\begin{equation}
\begin{aligned}
\left\|\nabla_{\theta^{(l)}} \right. & \left. \mathcal{L}(f_{\theta}; \mathcal{D}_i^q) - \nabla_{\theta^{(l)}} \mathcal{L}(f_{\theta}; \mathcal{D}_j^q)\right\|   = \left\|\sum_{k} 
\nabla_{\theta^{(l)}} \mathcal{L}\left(f_{\theta}\left(x_{ik} \right), y_{ik}\right) - \sum_{k} 
\nabla_{\theta^{(l)}} \mathcal{L}\left(f_{\theta}\left(x_{jk} \right), y_{jk}\right) \right\| \\
& = \left\| \sum_{k}  \left(\Delta_{ik}^{(l)} \Sigma_L^{\prime}\left(z_{ik}^{(L)}\right) \nabla^{(ik)}_{\theta^{(L)}} \mathcal{L}\right)\left(x_{ik}^{(l-1)}\right)^T  -  \sum_{k}  \left(\Delta_{jk}^{(l)} \Sigma_L^{\prime}\left(z_{jk}^{(L)}\right) \nabla^{(jk)}_{\theta^{(L)}} \mathcal{L}\right)\left(x_{jk}^{(l-1)}\right)^T  \right\| \\
& \leq \sum_{k} \left\{ \left\| \Delta_{ik}^{(l)} \right\| \cdot \left\|\left(x_{ik}^{(l-1)}\right)^T\right\| \cdot \left\| \Sigma_L^{\prime}\left(z_{ik}^{(L)}\right) \nabla^{(ik)}_{\theta^{(L)}} \mathcal{L}  -   \Sigma_L^{\prime}\left(z_{jk}^{(L)}\right) \nabla^{(jk)}_{\theta^{(L)}} \mathcal{L} \right\| \right.\\
& \left. \quad + \left\| \Sigma_L^{\prime}\left(z_{jk}^{(L)}\right) \nabla^{(jk)}_{\theta^{(L)}} \mathcal{L}  \right\| \cdot   \left\| \Delta_{ik}^{(l)}\left(x_{ik}^{(l-1)}\right)^T 
 -\Delta_{jk}^{(l)} \left(x_{jk}^{(l-1)}\right)^T\right\| \right\}\\
& \leq \sum_{k} \left\{ \left\| \Delta_{ik}^{(l)} \right\| \cdot \left\|\left(x_{ik}^{(l-1)}\right)^T\right\| \cdot \left\| \Sigma_L^{\prime}\left(z_{ik}^{(L)}\right) \nabla^{(ik)}_{\theta^{(L)}} \mathcal{L}  -   \Sigma_L^{\prime}\left(z_{jk}^{(L)}\right) \nabla^{(jk)}_{\theta^{(L)}} \mathcal{L} \right\| \right.\\
& \left. \quad + \left\| \Sigma_L^{\prime}\left(z_{jk}^{(L)}\right) \nabla^{(jk)}_{\theta^{(L)}} \mathcal{L}  \right\| \cdot  \left( \left\|\Delta_{ik}^{(l)}\right\| \cdot \left\|\left(x_{ik}^{(l-1)}\right)^T\right\|
 + \left\|\Delta_{jk}^{(l)}\right\| \cdot \left\|\left(x_{jk}^{(l-1)}\right)^T\right\| \right) \right\} \quad \text{By Triangle Inequality}\\
& \leq \sum_{k} \left\{  \max _{l, k}\left(\left\| \Delta_{ik}^{(l)} \right\| \cdot \left\|\left(x_{ik}^{(l-1)}\right)^T\right\| \right)\cdot\left\| \Sigma_L^{\prime}\left(z_{ik}^{(L)}\right) \nabla^{(ik)}_{\theta^{(L)}} \mathcal{L}  -   \Sigma_L^{\prime}\left(z_{jk}^{(L)}\right) \nabla^{(jk)}_{\theta^{(L)}} \mathcal{L} \right\|\right.\\
& \left. \quad +\left\| \Sigma_L^{\prime}\left(z_{jk}^{(L)}\right) \nabla^{(jk)}_{\theta^{(L)}} \mathcal{L}  \right\|  \cdot \max _{l, i, j}\left(\left\|\Delta_i^{(l)}\right\| \cdot\left\|x_i^{(l-1)}\right\|+\left\|\Delta_j^{(l)}\right\| \cdot\left\|x_j^{(l-1)}\right\|\right)\right\} \quad \text{Take Maximum over $l$,$k$,$i$,$j$}\\
& =  \underbrace{\max _{l, k}\left(\left\| \Delta_{ik}^{(l)} \right\| \cdot \left\|\left(x_{ik}^{(l-1)}\right)^T\right\| \right)}_{c_{1}} \cdot \sum_{k}\left\| \Sigma_L^{\prime}\left(z_{ik}^{(L)}\right) \nabla^{(ik)}_{\theta^{(L)}} \mathcal{L}  -   \Sigma_L^{\prime}\left(z_{jk}^{(L)}\right) \nabla^{(jk)}_{\theta^{(L)}} \mathcal{L} \right\|\\
&  \quad +\underbrace{\sum_{k}\left\| \Sigma_L^{\prime}\left(z_{jk}^{(L)}\right) \nabla^{(jk)}_{\theta^{(L)}} \mathcal{L}  \right\|  \cdot \max _{l, i, j}\left(\left\|\Delta_i^{(l)}\right\| \cdot\left\|x_i^{(l-1)}\right\|+\left\|\Delta_j^{(l)}\right\| \cdot\left\|x_j^{(l-1)}\right\|\right)}_{c_2}\\
\label{boundgrad}
\end{aligned}
\end{equation}

Thus, we derive the gradient of meta-model $f_{\theta} $ on $\mathcal{D}^{q}_{i}$ w.r.t.  the $l$-th layer $\theta^{(l)}$ can be bounded by the gradient of the loss w.r.t. the pre-activation outputs. $c_1$ and $c_2$ will be used for further derivation.

According to (\ref{boundgrad}), we can show that two arbitrary query sets' gradient of meta-model can be bounded by constant times the gradient of the loss w.r.t. the pre-activation outputs of the neural network as

\begin{equation}\label{boundgrad2}
    \left\|\nabla_{\theta} \mathcal{L}(f_{\theta}; \mathcal{D}_i^q) - \nabla_{\theta} \mathcal{L}(f_{\theta}; \mathcal{D}_j^q)\right\| \leq L\cdot c_{1} \left\| \underbrace{\sum_{k} \left(\Sigma_L^{\prime}\left(z_{ik}^{(L)}\right) \nabla^{(ik)}_{\theta^{(L)}} \mathcal{L}\right)}_{\Tilde{g}_{i}}  - \underbrace{\sum_{k} \left(  \Sigma_L^{\prime}\left(z_{jk}^{(L)}\right) \nabla^{(jk)}_{\theta^{(L)}} \mathcal{L}\right)}_{\Tilde{g}_{j}}  \right\| + L\cdot c_{2}
\end{equation}

Due to the bi-level optimization structure of meta-learning, the intrinsic gradient for outer loop meta-model updating is $\nabla_{\theta} \mathcal{L}(f_{\phi_{i}}; \mathcal{D}_i^q)$ for task $\mathcal{T}_{i}$. Suppose the loss function $\mathcal{L}$ is $\beta$-smooth, the norm of the outer loop gradient difference $\nabla_{\theta} \mathcal{L}(f_{\phi_{i}}; \mathcal{D}_i^q) - \nabla_{\theta} \mathcal{L}(f_{\phi_{j}}; \mathcal{D}_j^q)$ can be bounded based on the result of (\ref{boundgrad2}):

\begin{equation}
\begin{aligned}
& \left\|\nabla_{\theta} \mathcal{L}(f_{\phi_{i}}; \mathcal{D}_i^q) - \nabla_{\theta} \mathcal{L}(f_{\phi_{j}}; \mathcal{D}_j^q)\right\| \\
& =\left\|\nabla_{\theta} \mathcal{L}(f_{\phi_{i}}; \mathcal{D}_i^q) - \nabla_{\theta} \mathcal{L}(f_{\theta}; \mathcal{D}_i^q) + \nabla_{\theta} \mathcal{L}(f_{\theta}; \mathcal{D}_i^q) - \nabla_{\theta} \mathcal{L}(f_{\phi_{j}}, \mathcal{D}_j^q) + \nabla_{\theta} \mathcal{L}(f_{\theta}; \mathcal{D}_j^q) - \nabla_{\theta} \mathcal{L}(f_{\theta}; \mathcal{D}_j^q)\right\| \\
& \leq \left\|\nabla_{\theta} \mathcal{L}(f_{\phi_{i}}; \mathcal{D}_i^q) - \nabla_{\theta} \mathcal{L}(f_{\theta}; \mathcal{D}_i^q)\right\| + \left\|\nabla_{\theta} \mathcal{L}(f_{\phi_{j}}; \mathcal{D}_j^q) - \nabla_{\theta} \mathcal{L}(f_{\theta}; \mathcal{D}_j^q)\right\| + \left\|\nabla_{\theta} \mathcal{L}(f_{\theta}; \mathcal{D}_i^q) - \nabla_{\theta} \mathcal{L}(f_{\theta}; \mathcal{D}_j^q)\right\| \\
& = \underbrace{L \cdot c_{1}}_{C_{1}} \left\| \Tilde{g}_{i} - \Tilde{g}_{j}  \right\| + \underbrace{L \cdot c_{2} + \beta \cdot \left( \left\| \theta - \phi_i \right\| +\left\| \theta - \phi_j \right\| \right)}_{C_{2}} \\
\end{aligned}
\end{equation}

\section{Proof of Theorem 1} \label{theoprrof}
We follow the high-level idea of \cite{pooladzandi2022adaptive} and adapt the analysis to the bi-level updating setting of meta-learning.

The updated formula:
$$
\begin{aligned}
    &\theta^{t+1} = \theta^t - \eta\nabla \sum_{i\in S}\mathcal{L}(f_{\phi_i}; \mathcal{D}^q) \\
    &\text{where}\quad \phi_i = \theta^{t} - \eta'\nabla \sum_{(x,y) \in \mathcal{D}^s_i} \mathcal{L}(f_\theta ; \mathcal{D}^s_i)\\ 
\end{aligned}
$$
First, consider the task gradient:
\begin{equation}
    \label{gradient}
    \begin{aligned}
        \nabla_{\theta} \mathcal{L}(f_{\phi_i}; \mathcal{D}_i^q) &= \nabla_{\theta}\mathcal{L}(f_{\phi_i};\mathcal{D}^q_i) \\
        & = \nabla_{\theta}\phi_i\nabla_{\phi_i}\mathcal{L}(f_{\phi_i};\mathcal{D}_i^q) \\
        & = \nabla_{\theta}(\theta^t-\eta'\nabla_{\theta} \sum_{(x_i,y_i)\in \mathcal{D}_i^s}\mathcal{L}(f_{\theta};(x_i^s,y_i^s))\nabla_{\phi_i}\mathcal{L}(f_{\phi_i};\mathcal{D}_i^q)\\
        & = (-\eta'\sum_{(x_i,y_i)\in \mathcal{D}_i^s}\nabla^2_{\theta}\mathcal{L}(f_{\theta};(x_i^s,y_i^s))\nabla_{\phi_i}\mathcal{L}(f_{\phi_i};\mathcal{D}_i^q)
    \end{aligned}
\end{equation}
Similarly, by the $\beta$-smoothness (\textbf{Assumption 1}) of loss function $\mathcal{L}(f,\mathcal{D})$ and $\mu-PL^{\star}$ condition (\textbf{Assumption 2}), we can get:
\begin{equation}
    \begin{aligned}
    \mathcal{L}(\theta^{t+1}; & \mathcal{D}) - \mathcal{L}(\theta^{t};\mathcal{D}) \leq -\frac{\eta}{2}\| \nabla_{\theta}\sum_{i\in\mathcal{S}} \gamma_{j} \mathcal{L}(f_{\phi_i};\mathcal{D}_i^q)\|^2\\
    & \leq -\frac{\eta}{2}(\|\nabla_{\theta}\sum_{j\in\mathcal{M}}\mathcal{L}(f_{\phi_j};\mathcal{D}_j^q)\| - \epsilon)^2 \quad \text{Substitute weighted subset gradient as full gradient and $\epsilon$}\\ 
    & = -\frac{\eta}{2}(\|\nabla_{\theta}\sum_{j\in\mathcal{M}}\mathcal{L}(f_{\phi_j};\mathcal{D}_j^q)\|^2 -2\epsilon\|\nabla_{\theta}\sum_{j\in\mathcal{M}}\mathcal{L}(f_{\phi_j};\mathcal{D}_j^q)\| + \epsilon^2)\\
    &= -\frac{\eta}{2}(\|-\eta'\sum_{(x_j,y_j)\in \mathcal{D}_j^s}\nabla^2_{\theta}\mathcal{L}(f_{\phi_j};(x_j^s,y_j^s))\sum_{j\in\mathcal{M}}\nabla_{\phi_j}\mathcal{L}(f_{\phi_j};\mathcal{D}_j^q)\|^2 -2\epsilon\|\nabla_{\theta}\sum_{j\in\mathcal{M}}\mathcal{L}(f_{\phi_j};\mathcal{D}_j^q)\| + \epsilon^2) \quad \text{by \eqref{gradient}} \\
    &\leq -\frac{\eta\eta'm }{2}\|\sum_{j\in\mathcal{M}}\nabla_{\phi_j}\mathcal{L}(f_{\phi_j};\mathcal{D}_j^q)\|^2 + \eta \epsilon  \|\nabla_{\theta}\sum_{j\in\mathcal{M}}\mathcal{L}(f_{\phi_j};\mathcal{D}_j^q)\| - \frac{\eta}{2} \epsilon^2 \quad \text{ Bounded Hessian} \\
    &\leq -\eta\eta'm\mu \mathcal{L}(f_{\phi};\mathcal{D}^q) + \eta \epsilon \|\nabla_{\theta}\sum_{j\in\mathcal{M}}\mathcal{L}(f_{\phi_i};\mathcal{D}_i^q)\| - \frac{\eta}{2} \epsilon^2 \quad \text{By $\mu-PL^{\star}$ Condition}\\
    & = -\eta\eta'm\mu(\mathcal{L}(f_{\phi_i};\mathcal{D}) - \mathcal{L}(f_{\theta^t};\mathcal{D})) -\eta\eta'm\mu \mathcal{L}(f_{\theta^t};\mathcal{D})  + \eta \epsilon  \|\nabla_{\theta}\sum_{j\in\mathcal{M}}\mathcal{L}(f_{\phi_i};\mathcal{D}_i^q)\| - \frac{\eta}{2} \epsilon^2 \\
    & \leq -\eta\eta'm\mu\mathcal{L}(f_{\theta^t};\mathcal{D})  + \eta \epsilon \|\nabla_{\theta}\sum_{j\in\mathcal{M}}\mathcal{L}(f_{\phi_i};\mathcal{D}_i^q)\| - \frac{\eta}{2} \epsilon^2 + \eta\eta'm\mu|\mathcal{L}(f_{\phi*};\mathcal{D}) - \mathcal{L}(f_{\theta^0};\mathcal{D})|  \\
\end{aligned}
\end{equation}
Let $r = |\mathcal{L}(f_{\phi*};\mathcal{D}) - \mathcal{L}(f_{\theta^0},\mathcal{D})|$, we obtain: 
$$
\mathcal{L}(f_{\theta^{t+1}};\mathcal{D}) \leq (1 -\eta\eta'm\mu) \mathcal{L}(f_{\theta^{t}};\mathcal{D}) + \eta \epsilon 
 \|\nabla_{\theta}\sum_{j\in\mathcal{M}}\mathcal{L}(f_{\phi_i};\mathcal{D}_i^q)\| - \frac{\eta}{2} \epsilon^2 +  \eta\eta'm\mu r.
$$
This implies:
$$
\begin{aligned}
    \mathcal{L}(f_{\theta^{t}};\mathcal{D}) & \leq (1 -\eta\eta'm\mu)^t \mathcal{L}(f_{\theta^{0}};\mathcal{D}) + \sum_{k = 0}^{t-1}(1 -\eta\eta'm\mu)^k(\eta \epsilon \|\nabla_{\theta}\sum_{j\in\mathcal{M}}\mathcal{L}(f_{\phi_i};\mathcal{D}_i^q)\| - \frac{\eta}{2} \epsilon^2 + \eta\eta'm\mu r) \\
    &\leq  (1 -\eta\eta'm\mu)^t \mathcal{L}(f_{\theta^{0}};\mathcal{D}) + \frac{1}{\eta\eta'm\mu}(\eta \epsilon  \|\nabla_{\theta}\sum_{j\in\mathcal{M}}\mathcal{L}(f_{\phi_i};\mathcal{D}_i^q)\| - \frac{\eta}{2} \epsilon^2 + \eta\eta'm\mu r),
\end{aligned}
$$
giving:
$$
\mathbb{E}[\left(\mathcal{L}(f_{\theta^{t}};\mathcal{D})\right)] \leq (1 -\eta\eta'm\mu)^t \mathbb{E}[(\mathcal{L}(f_{\theta^{0}};\mathcal{D}))] + \frac{1}{\eta\eta'm\mu}(\eta \epsilon \|\mathbb{E}[\nabla_{\theta} \mathcal{L}(f_{\phi};\mathcal{D})]\|_{L^{\infty}_\mathcal{M}} - \frac{\eta}{2} \epsilon^2) +  r
$$
where
$$
    \|\mathbb{E}[\nabla_{\theta} \mathcal{L}(f_{\phi};\mathcal{D})]\|_{L^{\infty}_\mathcal{M}} =
    \sup_{\Gamma}\left\|\mathbb{E}\left[\sum_{j \in \mathcal{M}} \left( \nabla_{\theta}\mathcal{L}\left(f_{\phi_{j}} ; \mathcal{\mathcal{D}}_{j}^q\right)  -\nabla_{\theta}\mathcal{L}\left(f_{\phi_{\Gamma(j)}} ; \mathcal{\mathcal{D}}_{\Gamma(j)}^q \right)\right)\right]\right\|.
$$

\section{Implementation Details of Noisy Task Setting} \label{noise}

We provide the detailed noisy task-generating mechanism as follows.

\begin{algorithm}[h]

	\caption{Noisy Task Generating Mechanism (5-way 5-shot)}

	\begin{algorithmic}[1] 
		\REQUIRE Task $\mathcal{T}$; generating rate $\lambda$ ; noise threshold $t$
        \STATE Draw 5 samples ($\lambda_{1},\cdots,\lambda_{5}$) $\sim$ $Poisson(\lambda)$
        \FOR{$\lambda_{i}$}
            \IF{$\lambda_{i}$ $\geq t$}
                \STATE $\lambda_{i}$ = t
            \ENDIF
        \ENDFOR
        \FOR{class $i$}
        \STATE Randomly draw $\lambda_{i}$ samples ($c_{i,1},\cdots,c_{i,\lambda_{i}}$) from other $4$ classes 
            \FOR{$c_{i,j}$}
                \STATE Randomly draw data $x_{c_{i,j},k}$ from class $c_{i,j}$ and data $x_{i,k}$ from class $i$
                \STATE Switch the label of two selected data
            \ENDFOR
        \ENDFOR
        \STATE Randomly split task $\mathcal{T}_{i}$ into support set and query set
        \STATE Output the constructed noisy task

\end{algorithmic}
\label{alg:noise}
\end{algorithm}

\textbf{Algorithm \ref{alg:noise}} is based on symmetric label swap for few-shot learning \citep{liang2022few}. By the above label noise-generating mechanism, the mislabeled data could exist in both the support set and the query set. The noise ratio is controlled by noise threshold $t$ and $Poisson(\lambda)$ in \textbf{Algorithm \ref{alg:noise}}.

\section{Additional Experiment}\label{resnet}

\subsection{ ResNet-12 as Large Backbone}

To show DERTS works for a larger backbone, we explored the performance of ANIL and PN with ResNet-12 \citep{oreshkin2018tadam} in both limited data budget and noisy label task (noise ratio 40\%) settings on Mini-Imagenet. We keep the ResNet-12 configuration details the same as CNN4.

\begin{table*}[t]

\centering

\begin{tabular}{l|lll}
\hline \hline
Dataset & \multicolumn{3}{l}{Mini-ImageNet (5-way 5-shot)}\\
\hline
Iterations     & 10\%  & 30\% & All \\ \hline
ANIL     &  $52.33 \pm 0.7$    &  $62.75 \pm 0.8$    &   $68.01 \pm 0.7$       \\ 
ANIL-US        &   $56.08 \pm 0.7$   &   $58.92 \pm 0.7$   &    $\mathbf{68.34 \pm 0.6}$         \\ 
ANIL-ATS        &  $51.20 \pm 0.6$    &  $60.62 \pm 0.8$    &    $67.25 \pm 0.8$        \\ 
\textbf{ANIL-DERTS}      &   $\mathbf{56.24 \pm 0.7}$    &  $\mathbf{67.19 \pm 0.7}$    &  $68.23 \pm 0.6$  \\ \hline 
PN        &   $59.07 \pm 0.8$   &   $61.13 \pm 0.8$   &    $67.56 \pm 0.6$         \\ 
PN-US        &  $58.35 \pm 0.8$    &  $60.91 \pm 0.9$    &    $67.72 \pm 0.7$        \\ 
\textbf{PN-DERTS}      &   $\mathbf{60.17 \pm 0.9}$    &  $\mathbf{63.45 \pm 0.8}$    &  $\mathbf{67.81\pm 0.7}$  \\ \hline \hline

\end{tabular}
 
\caption{Average accuracy of 5-way 5-shot Mini-ImageNet Classification with Limited Data Budget Setting (ResNet-12 as the backbone). 10\% (30\%) in the table denotes the performance after learning on 10\% (30\%) tasks during the episodic training.} 
\label{res1}

\end{table*}

\begin{table*}[t]

\centering

\begin{tabular}{l|l}
\hline \hline
Dataset & Mini-ImageNet (5-way 5-shot)\\
\hline
Noise Ratio         & 40\%  \\ \hline
ANIL     &  $62.60 \pm 0.59$        \\ 
ANIL-US        &   $55.16 \pm 0.88$      \\ 
ANIL-ATS        &  $60.17 \pm 0.77$        \\ 
\textbf{ANIL-DERTS}      &   $\mathbf{64.87 \pm 0.72}$   \\ \hline
PN    &  $54.15 \pm 0.68$        \\ 
PN-US        &   $53.93 \pm 0.70$      \\ 
\textbf{PN-DERTS}      &   $\mathbf{55.43 \pm 0.65}$   \\ \hline \hline

\end{tabular}
 
\caption{Average accuracy of 5-way 5-shot Mini-ImageNe Classification with Noisy Task Setting (ResNet-12 as the backbone). 40\% in the table denote the noise ratio (percentage of mislabeled data).} 
\label{res2}

\end{table*}

From \textbf{Table \ref{res1}} and \textbf{\ref{res2}}, we observe that DERTS generally holds the advantage of data efficiency and robustness for both settings when shifting the backbone to ResNet-12. In the limited data budget setting, DERTS shows its faster learning capability towards baselines. In the noisy label task setting, DERTS for ANIL outperforms baseline by at least 3\% on accuracy, which significantly shows DERTS is effective for larger backbones. One thing worth mentioning here is that ANIL-US and PN-US perform comparably better on ResNet12 in the limited data budget setting than CNN4. We speculate the stronger representation capability of ResNet12 empowers ANIL-US and PN-US with a better ability to be aware of the difficulty of tasks, but it is still not robust in the noisy task setting compared to other methods.

\subsection{Additional Experiment on Limited Data budget}

In this subsection, we provide additional experiments and details for limited data budget setting. In the main context, we present the experiment results on only training 16 classes (25\% classes and 25\% data). Here, we provide experiment results for training 32 classes (50\% classes and 50\% data) in \textbf{Table \ref{50}}. According to the experiment results, DERTS outperforms all the baselines with an average of 1.2\% in accuracy, which further indicates that DERTS captures the task diversity in this setting with fewer training classes.

According to \cite{yao2021meta}, the details of selected training classes are as follows. For 25\% training classes (16 classes), we select: $\{\mathrm{n} 02823428, \mathrm{n} 13133613, \mathrm{n} 04067472, \mathrm{n} 03476684$, n02795169, n04435653, n03998194, n02457408, n03220513, n03207743, n04596742, n03527444, n01532829, n02687172, n03017168, n04251144\}.

In addition, the selected classes for 50\% training classes are: $\{\mathrm{n} 03676483, \mathrm{n} 13054560, \mathrm{n} 04596742, \mathrm{n} 01843383$, n02091831, n03924679, n01558993, n01910747, n01704323, n01532829, n03047690, n04604644, n02108089, n02747177, n02111277, n01749939, n03476684, n04389033, n07697537, n02105505, n02113712, n03527444, n03347037, n02165456, n02120079, n04067472, n02687172, n03998194, n03062245, n07747607, n09246464, n03838899 \}.

\begin{table}[t!]
\vspace{-0.2cm}
\centering

\begin{tabular}{l|ll}
\hline \hline
Dataset & \multicolumn{2}{l}{Mini-ImageNet (32-Class)} \\
\hline
Method          & 5-way 5-shot & 5-way 1-shot\\ \hline
ANIL     &  $54.86 \pm 0.72$    &  $42.85 \pm 0.71$    \\ 
ANIL-US        &   $55.07 \pm 0.70$   &   $42.72 \pm 0.60$         \\ 
ANIL-ATS        &  $54.61 \pm 0.68$    &  $42.55 \pm 0.64$       \\ 
\textbf{ANIL-DERTS}      &   $\mathbf{56.10 \pm 0.64}$    &  $\mathbf{43.97 \pm 0.58}$   \\ \hline 
ProtoNet        &   $59.29 \pm 0.69$   &   $43.17\pm 0.61$         \\ 
ProtoNet-US        &  $59.21 \pm 0.61$    &  $42.75\pm 0.72$       \\ 
\textbf{ProtoNet-DERTS}      &   $\mathbf{60.17\pm 0.68}$    &  $\mathbf{44.20\pm 0.71}$   \\ \hline \hline
\end{tabular}
\caption{5-way 5-shot / 5-way 1-shot Mini-ImageNet Classification with 50\% Class Training Set.}
\label{50}
\vspace{-0.3cm}
\end{table}

\subsection{Case Study for Task Selection}

\begin{figure}[ht!]
     \centering
     \vspace{-0.3cm}
    \includegraphics[scale=0.48]{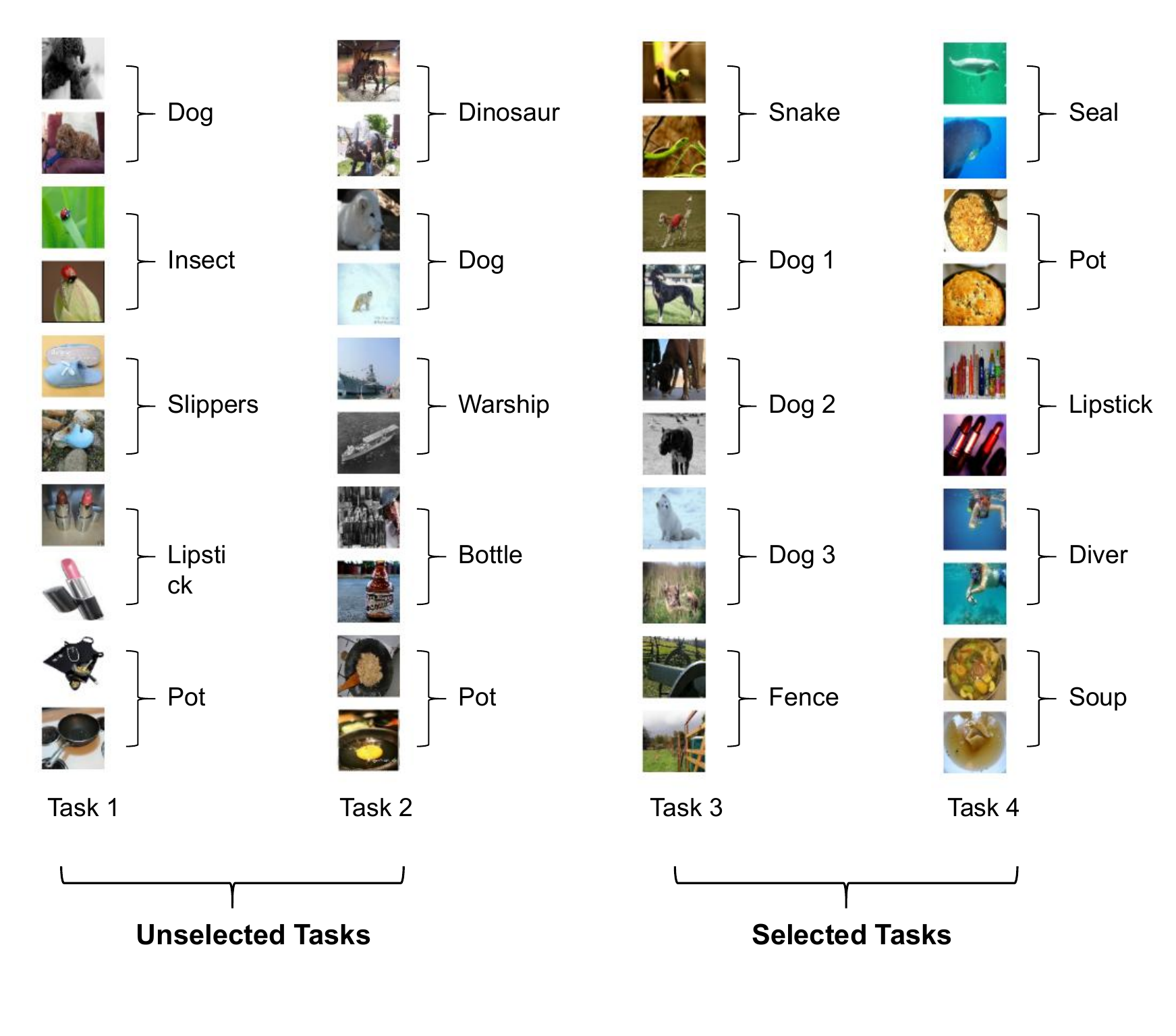}
     \vspace{-0.8cm}
     \caption{Typical examples of selected tasks by DERTS and unselected tasks.}
    \label{case}
    \vspace{-0.5cm}
\end{figure}

We provide a brief case study for further analysis of the tasks selected and not selected by DERTS. \textbf{Figure \ref{case}} displays typical examples of both selected and unselected tasks. From the two unselected tasks presented, we observe that these tasks are generally coarse-grained. The classes within the unselected tasks are easily distinguishable, indicating that they might be relatively simpler. In contrast, the classes and images in the selected tasks tend to be more visually confusing. Task 3 includes three different classes of dogs, making this task more fine-grained than some of the unselected tasks. Task 4 consists of two visually similar pairs: the Seal and Diver pair, which might share the same background, and the Pot and Soup pair, which could have similar shapes and colors. The selected subset of tasks likely offers better diversity and is more challenging to learn, making them more informative for the meta-training process. This observation aligns with the claim made by related works on task sampling \citep{liu2020adaptive, yao2021meta, arnold2021uniform}.

\end{document}

%% file: preamble.tex
\usepackage[dvipsnames]{xcolor}
